\documentclass{article}

\usepackage[final]{tackling_climate_workshop_style}

\usepackage[utf8]{inputenc} 
\usepackage[T1]{fontenc}    
\usepackage{hyperref}       
\usepackage{url}            
\usepackage{booktabs}       
\usepackage{amsfonts}       
\usepackage{nicefrac}       
\usepackage{microtype}      
\usepackage{graphicx}
\usepackage[table]{xcolor}
\usepackage{caption, makecell, floatrow, soul}
\usepackage{xcolor}
\newcolumntype{L}[1]{>{\raggedright\arraybackslash}m{#1}}
\newcolumntype{C}[1]{>{\centering\arraybackslash}m{#1}}
\newcolumntype{R}[1]{>{\raggedleft\arraybackslash}m{#1}}

\newcommand{\suggestion}[1]{\textcolor{purple}}

\title{Global 3D Reconstruction of \\ Clouds \& Tropical Cyclones} 

\author{%
  Shirin Ermis \\
  University of Oxford \\
  \texttt{shirin.ermis@physics.ox.ac.uk} 
  \And
  Cesar Aybar \\
  Universitat de València \\
  \texttt{cesar.aybar@uv.es} 
  \And
  Lilli Freischem \\
  University of Oxford \\
  \texttt{lilli.freischem@physics.ox.ac.uk} 
  \And
  Stella Girtsou\\
  National Observatory of Athens \\
  National Technical University of Athens \\
  \texttt{girtsou.s@gmail.com} 
  \And
  Kyriaki-Margarita Bintsi \\ 
  Harvard Medical School and \\
  Massachusetts General Hospital \\
  \texttt{kbintsi@mgh.harvard.edu}
  \And
  Emiliano Diaz Salas-Porras\\
  Universitat de València \\
  \texttt{emdiazsal@gmail.com} 
  \And
  Michael Eisinger\\
  European Space Agency \\
  \texttt{michael.eisinger@esa.int} 
  \And
  William Jones \\
  University of Oxford \\
  \texttt{william.jones@physics.ox.ac.uk} 
  \And
  Anna Jungbluth \\
  European Space Agency \\
  \texttt{anna.jungbluth@esa.int} 
  \And
  Benoit Tremblay \\
  Environment and Climate Change Canada \\
  \texttt{benoit.tremblay@ec.gc.ca} 
}

\author{
\begin{minipage}[t]{0.45\textwidth}
\centering
\textbf{Shirin Ermis} \\
\normalfont University of Oxford \\
\texttt{shirin.ermis@physics.ox.ac.uk} \\[3.5ex]
\textbf{Lilli Freischem} \\
\normalfont University of Oxford \\
\texttt{lilli.freischem@physics.ox.ac.uk} \\
\texttt{ }\\[3.5ex]
\textbf{Kyriaki-Margarita Bintsi} \\
\normalfont Harvard Medical School and \\
\normalfont Massachusetts General Hospital \\
\texttt{kbintsi@mgh.harvard.edu} \\[3.5ex]
\textbf{Michael Eisinger} \\
\normalfont European Space Agency \\
\texttt{michael.eisinger@esa.int} \\[3.5ex]
\textbf{Anna Jungbluth} \\
\normalfont European Space Agency \\
\texttt{anna.jungbluth@esa.int}
\end{minipage}
\hfill
\begin{minipage}[t]{0.55\textwidth}
\centering
\textbf{Cesar Aybar} \\
\normalfont Universitat de València \\
\texttt{cesar.aybar@uv.es} \\[3.5ex]
\textbf{Stella Girtsou} \\
\normalfont National Observatory of Athens \\
\normalfont National Technical University of Athens \\
\texttt{girtsou.s@gmail.com} \\[3.5ex]
\textbf{Emiliano Diaz Salas-Porras} \\
\normalfont Universitat de València \\
\texttt{emdiazsal@gmail.com} \\
\texttt{ }\\[3.5ex]
\textbf{William Jones} \\
\normalfont University of Oxford \\
\texttt{william.jones@physics.ox.ac.uk} \\[3.5ex]
\textbf{Benoit Tremblay} \\
\normalfont Environment \& Climate Change Canada \\
\texttt{benoit.tremblay@ec.gc.ca}
\end{minipage}
}

\begin{document}

\maketitle

\begin{abstract}
    Accurate forecasting of tropical cyclones (TCs) remains challenging due to limited satellite observations probing TC structure and difficulties in resolving cloud properties involved in TC intensification.
    Recent research has demonstrated the capabilities of machine learning methods for 3D cloud reconstruction from satellite observations. 
    However, existing approaches have been restricted to regions where TCs are uncommon, and are poorly validated for intense storms.
    We introduce a new framework, based on a pre-training--fine-tuning pipeline, that learns from multiple satellites with global coverage to translate 2D satellite imagery into 3D cloud maps of relevant cloud properties. We apply our model to a custom-built TC dataset to evaluate performance in the most challenging and relevant conditions. We show that we can -- for the first time -- create global instantaneous 3D cloud maps and accurately reconstruct the 3D structure of intense storms. Our model not only extends available satellite observations but also provides estimates when observations are missing entirely. This is crucial for advancing our understanding of TC intensification and improving forecasts.

\end{abstract}

\section{Introduction}

Tropical cyclones (TCs) are the most damaging and costly extreme events worldwide, with damages reaching billions of dollars per storm in the United States alone \citep{noaa_us_2023}. Although numerical and machine learning (ML) weather forecasts for TCs have improved in recent decades, significant challenges remain to accurately forecast their paths and intensities \citep{elsberry_new_2025}. Rapid intensification, a process in which TC winds increase by more than 30 knots in 24 hours, is particularly difficult to forecast \citep{leroux_recent_2018, hendricks_summary_2019} and occurs in the majority of the most intense and damaging TCs \citep{sippel_tropical_2015}. 
Recent studies have shown that cloud microphysics in TCs can play an important role in their intensification \citep{ruppert_critical_2020}, as the vertical structure of ice clouds causes radiative heating of the clouds and drives instability \citep{wing_acceleration_2022}. 
Observational studies of the influence of clouds on TC intensification \citep{lee_satellite-based_2024} use the cloud profiling radar (CPR) aboard NASA's CloudSat mission \citep{stephens_cloudsat_2002}, which measures the vertical distribution and structure of clouds and their microphysical properties \citep{stephens_cloudsat_2018}.
However, CloudSat is limited by its long revisit time ($\sim$16 days), narrow swath (1.4 km), and fixed time of day sampling. This also applies to lidar instruments (e.g., onboard NASA's  CALIPSO \citep{winker_calipso_2010}) and novel satellites like ESA's EarthCARE mission \citep{earthCARE}. In contrast, geostationary imagery provides large coverage, typically every 10 minutes, but is restricted to measurements of cloud tops. 

Deriving information in the vertical domain from observations of cloud tops is challenging, but can be achieved - for instance - through cloud tomography techniques like stereo photogrammetry \citep{Horvth2001, Volkmer2024}. This involves collecting multiple perspectives of the same scene and using classical computer vision techniques to estimate depth and derive, for example, cloud top height, as is operationally done for NASA's MISR mission \cite{Horvth2001}. However, simultaneous observations from multiple perspectives are rarely available for satellites, and classical cloud tomography provides limited information below cloud tops. 
Beyond photogrammetry, vertical information can be derived by combining aligned observations from imaging and profiling sensors. Motivated by the recent launch of ESA's EarthCARE mission, Barker et al. developed a statistical pattern-matching algorithm to operationally extend observations from narrow vertical profiles and provide estimates of cloud volumes \cite{Barker2011}.
Following on this, recent research has demonstrated successes in the application of ML models to predict 3D volumes of CloudSat measurements from geostationary satellite imagery, including the use of U-Nets \citep{ronneberger_unet_2015} for predicting radar reflectivity (Z) \citep{bruning_artificial_2024}, ice water content (IWC) and ice crystal number concentration \citep{jeggle_icecloudnet_2024a}, and the use of vision transformers \citep{dosovitskiy_image_2021, girtsou_3d_2025}. ML is particularly suitable for this task, as it can effectively learn intricate spatial, temporal, and spectral patterns from large datasets.
Building on these advances, we introduce a novel pre-training--fine-tuning framework that integrates data from multiple geostationary satellites to enable global, near real-time 3D prediction of key cloud and TC properties including Z, IWC, and droplet effective radius ($\mathrm{r_e}$), facilitating detailed analysis of TC structure at high temporal cadence. Our model is based on a SWinMAE architecture \citep{xu2023swin}, and encodes temporal and spatial context, including solar and satellite viewing geometry. We propose the first geospatially-aware ML model for global, near real-time 3D cloud reconstructions and 3D reconstruction of tropical cyclones.

\section{Data} 

We compiled new multi-sensor datasets for 3D reconstruction of cloud/TC structure and microphysics, combining imagery from three geostationary satellites to achieve unprecedented diversity in viewing angles, cloud types, and geographic coverage, with co-located vertical profiles from CloudSat.

\textbf{Geostationary satellite imagery. } We use reflectance and brightness temperature (BT) data from three geostationary satellites: Meteosat Second Generation (MSG)/SEVIRI (centered at $0^\circ$ longitude, 11 spectral channels, 3 km resolution at nadir, from 2004) \citep{aminou_msg_2002}, Himawari-8/AHI (centered at $140.7^\circ$ E, 16 spectral channels, 2 km resolution at nadir, from 2015) \citep{bessho_introduction_2016}, and NOAA’s GOES-16/ABI (centered at $75.2^\circ$ W, 16 spectral channels, 2 km resolution at nadir, from 2018) \citep{schmit_closer_2016}. Each geostationary imager has a field of view of $\pm 80^\circ$ which we limit to $\pm 45^\circ$ to reduce distortion effects. Full-disk scans are taken every 10 minutes by GOES-16 and Himawari, and every 15 minutes by MSG, enabling continuous monitoring of clouds and TCs.

\textbf{Vertical profiles. } We use vertical profiles of Z \citep{marchand_hydrometeor_2008}, IWC, and $\mathrm{r_e}$ \citep{deng_cloudsat_2015} retrieved from CloudSat's CPR \citep{stephens_cloudsat_2002} and CALIPSO's lidar \citep{winker_calipso_2010}. The full data record from 2006 to 2020 is used (daytime only from 2012 after CloudSat's battery failure).

\textbf{ML-ready datasets. } We prepared three ML-ready datasets from the geostationary satellite imagery and vertical profiles: (1) a \textit{pre-training dataset} consisting of 50,000 randomly sampled patches of $1024 \times 1024$ pixels per satellite; (2) a \textit{clouds dataset} consisting of geostationary satellite imagery and spatially-temporally aligned CloudSat overpasses, and (3) a dedicated \textit{TC dataset} consisting of imagery and overpasses over tropical cyclones. For each image-profile pair, we align profiles of Z, IWC, $\mathrm{r_e}$ and cloud type classification through nearest neighbour averaging to the closest geostationary sensor pixel. More details on our datasets can be found in appendix tables \ref{appendix-table-pre-training-dataset}, \ref{appendix-table-finetuning-dataset}, \ref{appendix-table-TC-dataset}.




\section{Method}

The objective of our model is to translate 2D multi-spectral imagery from geostationary satellites into 3D volumes of cloud properties. For each image-profile pair, metrics (including the model loss) are calculated only over the narrow ground-truth CloudSat measurement. 

\textbf{Data normalisation. } To combine the diverse geostationary satellite sensors and create a unified input to our model, the 11 spectral channels with wavelengths closest to those of MSG/SEVIRI are selected from GOES/ABI and Himawari/AHI. Each spectral channel is normalised to a range of [-1, 1] using min-max normalisation (reflectances: 0--100\%; BT: 180--350\,K). The target CloudSat vertical profiles are normalised to [-1, 1] using min-max normalisation (Z: -30--20\,dBz; IWC: $10^{-5}$--10\,$\mathrm{gm^{-3}}$; $\mathrm{r_e}$: 0--160\,$\mathrm{\mu m}$). IWC is log-normalised to account for the large skew in its distribution. From the original 125 vertical height levels in the CloudSat data, we remove the lowest 20 levels (below ground level) and upper 25 levels (above cloud level), leaving 80 height levels.

\textbf{Baseline. } We compare our model to a state-of-the-art approach for 3D cloud reconstruction \citep{bruning_artificial_2024}. This approach uses a 2D residual U-Net \citep{diakogiannis_resuneta_2020} of depth 4, with 32 channels in the initial convolution layer. The output layer produces 80 height levels.

\textbf{Pre-training. } A SWin transformer–based \citep{liu_swin_2021} masked autoencoder \citep[SWinMAE:][]{xu2023swin} is used for large-scale self-supervised pre-training on unlabelled geostationary imagery. Since CloudSat measures vertical profiles via a sun-synchronous orbit (i.e. measuring each location always at the same local time), our target data is inherently temporally biased. Pre-training on general cloud scenes sampled outside of CloudSat overpasses helps to overcome this constraint.
During pre-training, we mask 50\% of the input image, tasking the model to use spatial context to reconstruct missing information. Examples of image reconstructions are shown in appendix fig. \ref{appendix-figure-pre-training}. The SWin transformer backbone offers two main advantages over a traditional vision transformer: hierarchical feature extraction and computational efficiency. Combined, this enables the model to capture fine local structures and global mesoscale cloud organization. In each training step, we randomly crop $256 \times 256$ pixel patches from our larger $1024 \times 1024$ pre-training dataset. Each batch is satellite-consistent, but up to all three geostationary satellites are shown to the model during training. We compare two configurations: (i) spectral-only input, and (ii) spectral plus metadata embeddings (SWinSatMAE: time, coordinates, solar/satellite viewing angles) combining elements from SatMAE \citep{cong_satmae_2023}. More training details can be found in appendix section \ref{appendix-training-details}.

\begin{figure}[htb]
    \centering
    \includegraphics[width=0.99\linewidth, trim=0 0 0 0, clip]{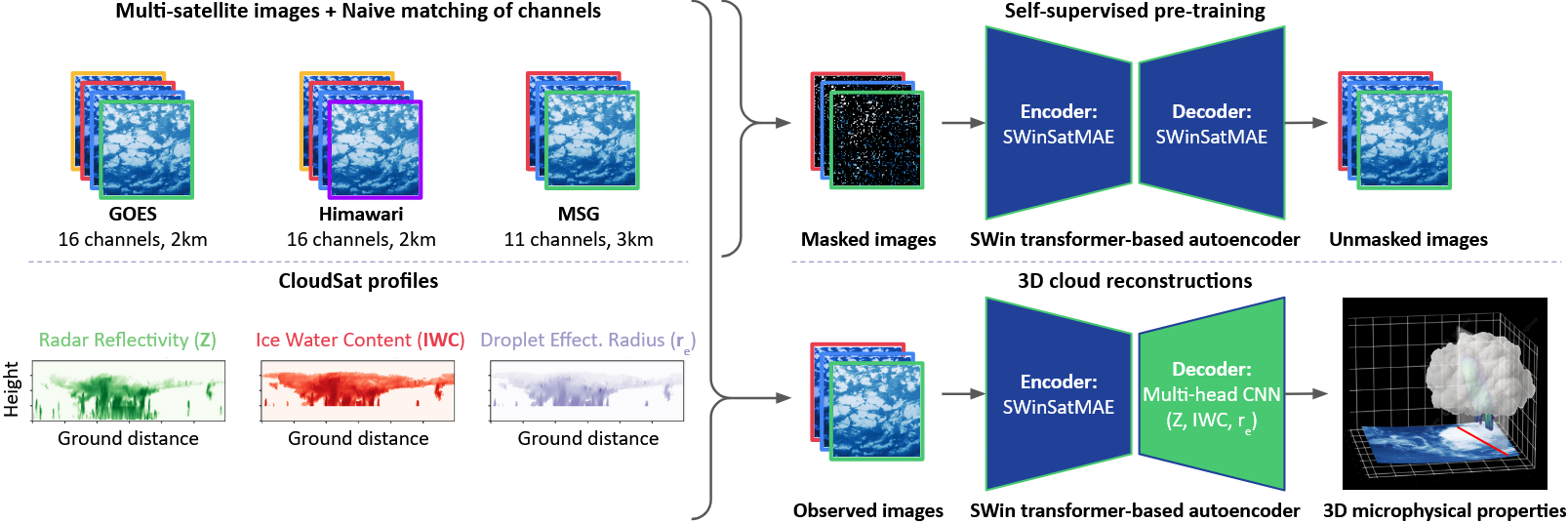}
    \caption{Overview of our ML pipeline. We select the 11 closest-matched spectral channels from the 16 channels of GOES and Himawari to create a consistent model input. During pre-training, the image encoder and decoder learn cloud structures by reconstructing masked images. During fine-tuning a 3D decoder is trained using paired image-profile pairs. Multiple prediction heads are used to predict different cloud properties simultaneously.}
    \label{fig:MLPipeline}
\end{figure}

\textbf{Fine-tuning. } For fine-tuning, we replace the MAE image reconstruction head with a task-specific 3D convolutional decoder that outputs a cloud property volume. We can perform either single-variable or multi-variable predictions using multiple output heads. In the latter case, we train the model to output Z, IWC, and $\mathrm{r_e}$ simultaneously, leveraging the shared structure and cross-correlation between these variables for improved predictions. An overview of our model pipeline is shown in fig. \ref{fig:MLPipeline}.

\section{Results} 

\begin{figure}[H]
    \vspace{-0.2cm}
    \centering
    \includegraphics[width=\linewidth, trim=0 0 0 3, clip]{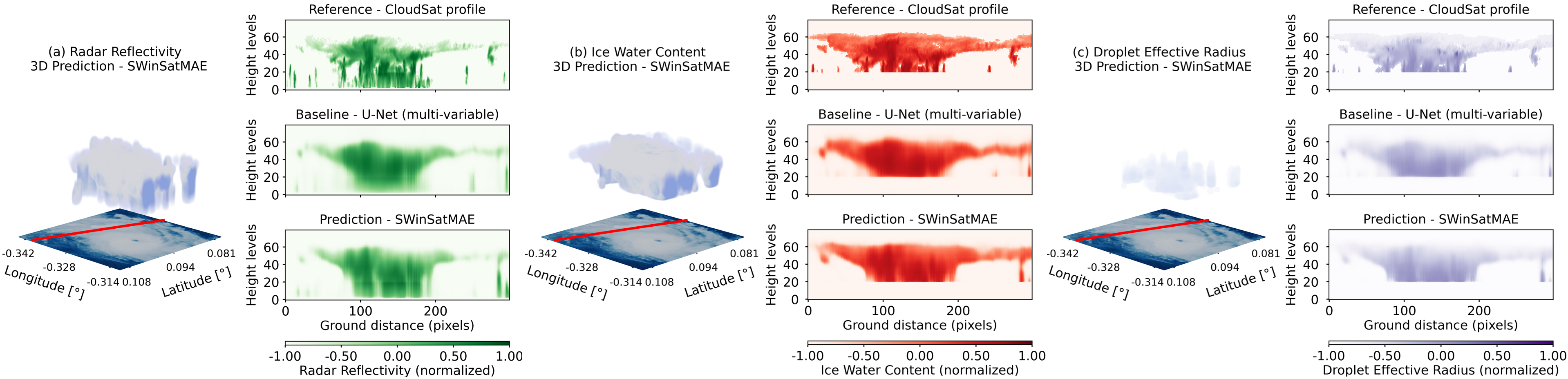}
    \caption{3D reconstructions of (a) Z, (b) IWC, and (c) $\mathrm{r_e}$ by the SWinSatMAE model for TC Dorian. The geostationary image from GOES channel 7 is shown under each 3D render, with the location of the CloudSat track marked in red. For validation purposes, the SWinSatMAE predictions along the CloudSat overpass are compared to the CloudSat retrievals and to the multi-variable U-Net baseline.}
    \label{fig:cross-sections}
\end{figure}
\begin{figure}[H]
    \vspace{-0.435cm}
    \centering
    \includegraphics[width=\linewidth, trim=0 8 0 8, clip]{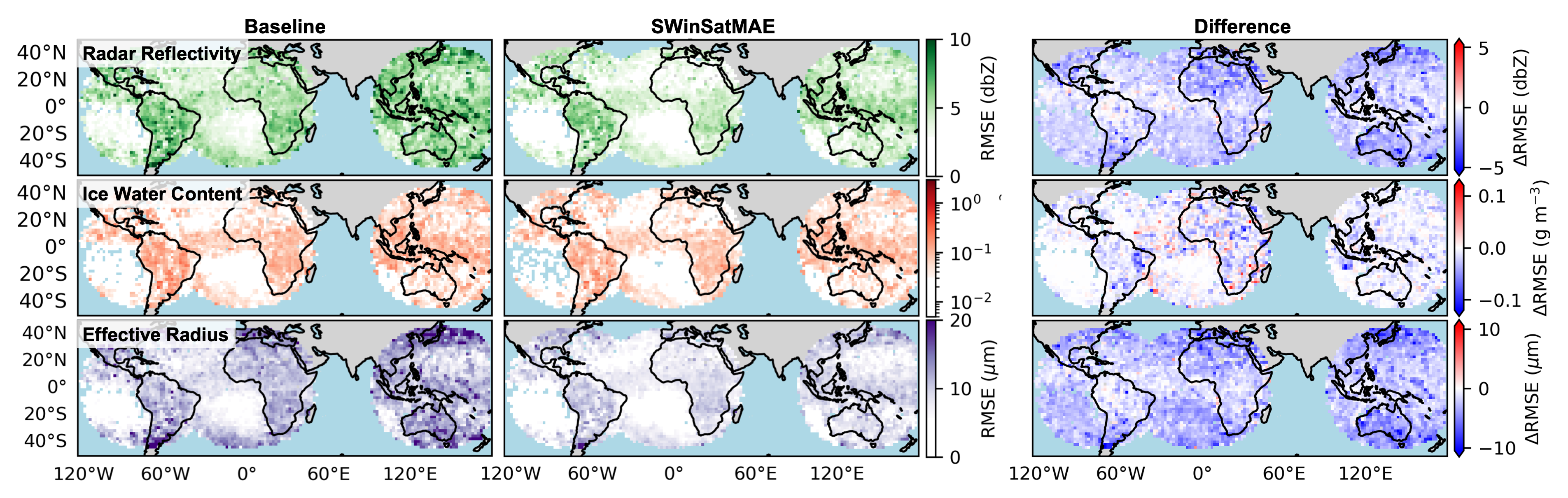}
    \caption{Spatial RMSE distribution of Z (top), IWC (middle), and $\mathrm{r_e}$ (bottom) predictions for the baseline model (left) and the SWinSatMAE model (middle), along with their difference (right).}
    \label{fig:spatial-rmse-comparison}
\end{figure}

\textbf{Baseline comparison. } In comparison to the U-Net, our model performs better at predicting Z, IWC, and $\mathrm{r_e}$ with lower root-mean-squared-error (RMSE) for all variables for both general cloud scenes and TCs (see appendix tables \ref{appendix-table-comparison-all}, \ref{appendix-table-comparison-cloudy}). For TCs, the SWinSatMAE produces more accurate values, and better predicts cloud top and base height (fig.~\ref{fig:cross-sections}). Spatially, the SWinSatMAE produces more consistent predictions, with improvements over land and at higher satellite viewing angles (fig.~\ref{fig:spatial-rmse-comparison}).

\textbf{Single vs multi-satellite:} To evaluate the effects of including different sensors in our model training, we compare the baseline model to three U-Nets trained on each geostationary satellite separately. While the multi-satellite model has higher RMSE for MSG, it has lower RMSE for GOES and Himawari, indicating that the larger training dataset helps improve performance for these sensors, and the simple channel matching approach does not significantly affect model performance (see appendix table~\ref{appendix-table-singlesat-multisat}). The multi-satellite model improves predictions for TC regions.

\textbf{Single vs multi-variable. } U-Net models trained to predict a single variable each are compared to the baseline model. Overall, the multi-variable model produces better predictions across all variables for all metrics. For TCs, the multi-variable model has lower RMSE for Z and $\mathrm{r_e}$ (see appendix table~\ref{appendix-table-singlevar-multivar}).

\textbf{Pre-training \& encoding. } We compare the pre-trained SWinSatMAE model to a SWinSatMAE trained from scratch in fine-tuning and a SwinMAE model without meta-data encoding. Overall, the pre-trained SWinSatMAE produces lower RMSE across all variables. For TCs, the SwinMAE produces better predictions, but the difference to the pre-trained SWinSatMAE is small (see appendix tables~\ref{appendix-table-comparison-all} and \ref{appendix-table-comparison-cloudy}). When breaking down the metrics by cloud type, the SWinSatMAE performs better than the other models in all cloudy conditions for $\mathrm{r_e}$, and in most cloud types for Z and IWC, but has worse metrics in clear skies (see appendix tables~\ref{appendix-table-comparison-all}, \ref{appendix-table-comparison-cloudy}, and \ref{appendix-table-cloud-types-comparison}).

\section{Conclusion}

\begin{figure}[htb]
    \centering    \includegraphics[width=0.9\linewidth]{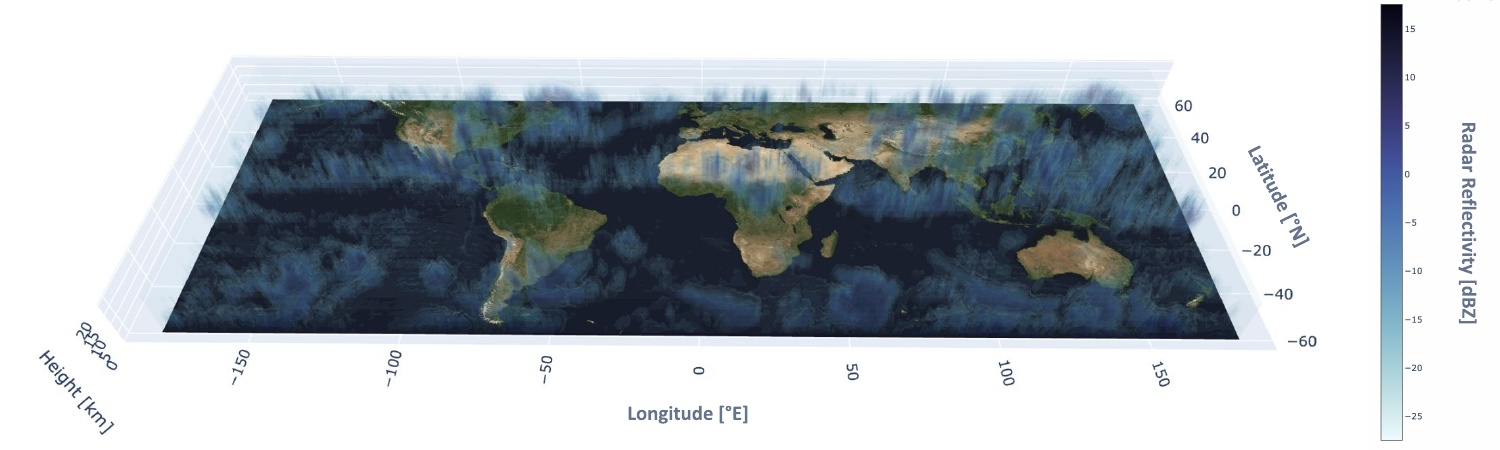}
    \caption{Using our model, we can --for the first time-- generate global instantaneous 3D cloud maps.}
    \label{appendix-figure-global-cloud-map}
\end{figure}

\begin{figure}[htb]
    \centering    \includegraphics[width=0.9\linewidth]{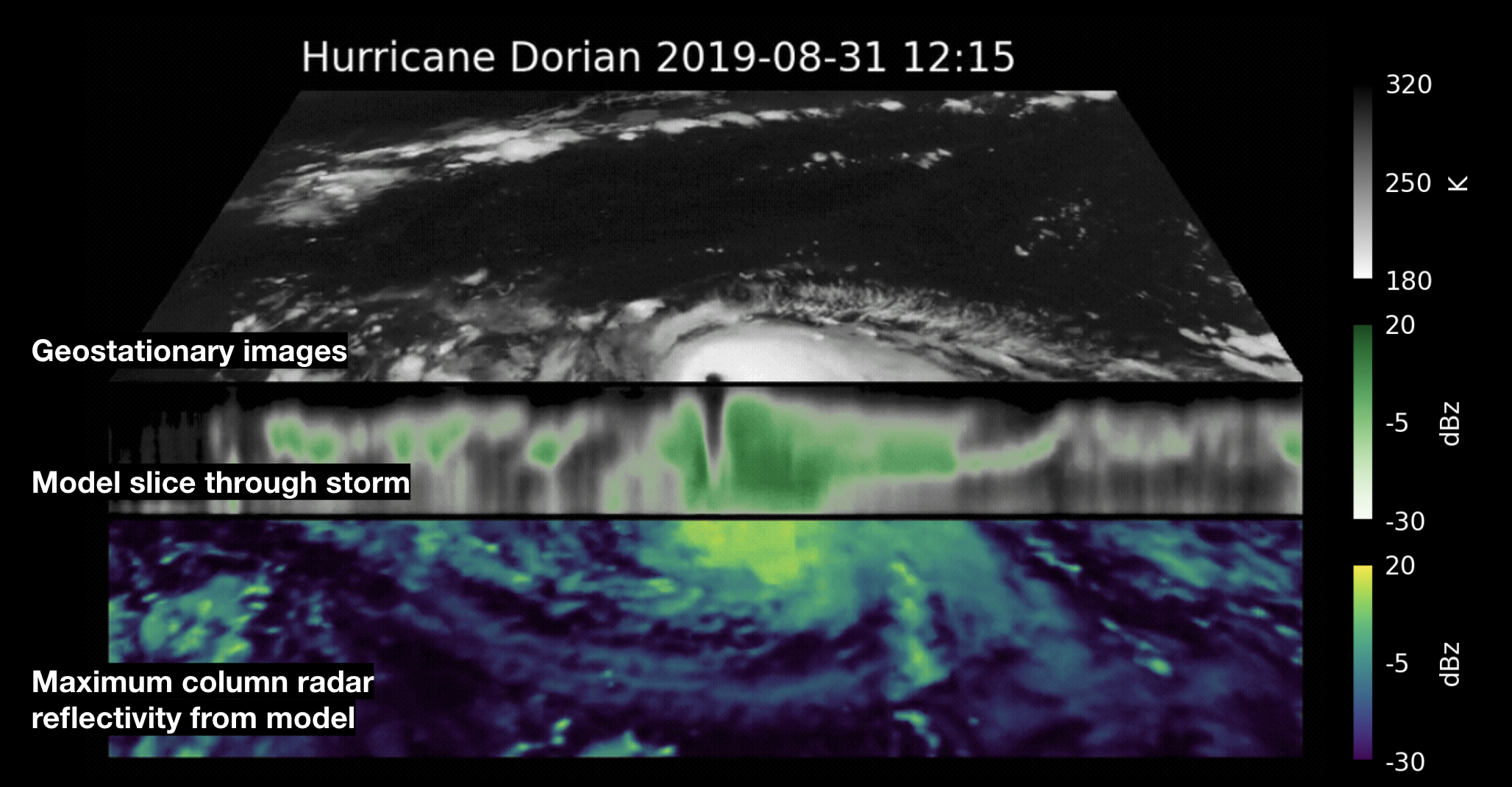}
    \caption{Model prediction for tropical cyclone Dorian. The top shows a geostationary satellite image, middle our model reconstruction of radar reflectivity, and bottom the max column radar reflectivity of the 3D prediction.}
    \label{appendix-figure-dorian-image-prediction}
\end{figure}

Measurements of vertically-resolved cloud properties are essential for understanding the complex processes involved in TC intensification. 
To address this need, we curated an AI-ready dataset consisting of pairs of geostationary satellites images and CloudSat overpasses, including a TC-specific subset. 
Using this dataset, we developed and trained the SWinSatMAE model; an architecture inspired by SWin transformers and masked autoencoders, which advances the state-of-the-art in 3D cloud reconstruction by predicting multiple microphysical properties with higher accuracy than existing approaches. Beyond improvements in accuracy, the model leverages multiple geostationary satellites to enable near real-time, global 3D predictions of clouds (appendix fig. \ref{appendix-figure-global-cloud-map}) and TCs (appendix fig. \ref{appendix-figure-dorian-image-prediction}), even with limited paired observational data. 
In effect, the trained model enhances the observational capabilities of available geostationary satellites by enabling the inference of microphysical properties from the radiance measurements as if a virtual CloudSat were observing the same field of view. 

While our model shows improvements already, there are certain areas where future research could lead to further advancement: 

\textbf{Error characterization.} The model tends to smooth cloud edges, reflecting high uncertainty in these regions. Several loss functions—including Huber, total variation, and Gaussian mixture loss—were tested without yielding a definitive improvement. Future work should examine probabilistic or generative methods capable of sharpening cloud boundaries while maintaining overall accuracy. Current error characterizations are limited to region and cloud type; analyses by altitude, climatological regime, and local weather conditions can be carried out and inform future model improvements.


\textbf{Validation.} Additionally, a more comprehensive validation is essential before this model can be applied to downstream scientific or operational tasks. This requires comparison against independent sources of information. We identify three main categories: (i) airborne field campaigns providing in situ and remote-sensing measurements, (ii) ground-based remote-sensing networks, and (iii) satellite-based observations.
Airborne campaigns such as ORCESTRA \citep{orcestra} provide in situ sampling of cloud microphysical properties, together with airborne radar, lidar, and polarimetric observations. These measurements can directly constrain estimates of reflectivity, bulk water content, and effective radius.
Ground-based networks, such as the Atmospheric Radiation Measurement program \citep[ARM: ][]{ARM}, deliver continuous datasets that include radar reflectivity profiles, ice and liquid water content retrievals from radar–lidar synergy, and effective radius estimates from combined radar–lidar–microwave approaches. 
Finally, satellite observations from EarthCARE \citep{earthCARE} provide significant advantages over CloudSat, including improved lidar vertical resolution, the addition of Doppler velocities, and collocated broadband radiometry. Together, these enable more accurate and vertically resolved microphysical retrievals, as well as direct evaluation of the radiative consistency of model outputs.

\textbf{Sensor dependence.} A fully sensor-independent approach \citep[e.g.,][]{francis_sensor_2024} could improve multi-satellite generalization and exploit the full spectral range of modern sensors. Furthermore, expanded validation on TCs will be critical to improve reliability in the most challenging storm conditions.
While numerical and ML forecasting models capture large-scale TC dynamics, they struggle to resolve the cloud processes most closely tied to rapid intensification. Satellites such as CloudSat offer valuable vertical information but lack the temporal resolution required for operational forecasting. By enabling near real-time 3D predictions of TC cloud properties from geostationary observations, our approach complements existing forecasting systems and offers new opportunities to better anticipate and mitigate the impacts of these devastating weather events. 

\begin{ack}
This work has been enabled by Frontier Development Lab Earth Systems Lab (https://eslab.ai/)---a public / private partnership between the European Space Agency (ESA), Trillium Technologies, the University of Oxford and leaders in commercial AI supported by Google Cloud, Scan Computers, Nvidia Corporation and Pasteur Labs.
The authors would also like to thank the reviewers and experts that provide advice throughout this project, including Julien Boussard, Gherardo Varando, Homer Durand, Milton Gomez, Johanna Mayer, Luis Gómez-Chova, Matteo Salvador, Alistair Francis, Mikolaj Czerkawski, Jacqueline Campbell, Emmanuel Johnson, Howard Barker, Sarah Brüning, Arthur Avenas, Dominique Brunet, Loredana Spezzi and Alessio Bozzo.
\end{ack}

\clearpage

\bibliographystyle{unsrt}  
\bibliography{references}

@misc{cong_satmae_2023,
  title = {{{SatMAE}}: {{Pre-training Transformers}} for {{Temporal}} and {{Multi-Spectral Satellite Imagery}}},
  shorttitle = {{{SatMAE}}},
  author = {Cong, Yezhen and Khanna, Samar and Meng, Chenlin and Liu, Patrick and Rozi, Erik and He, Yutong and Burke, Marshall and Lobell, David B. and Ermon, Stefano},
  year = {2023},
  month = jan,
  number = {arXiv:2207.08051},
  eprint = {2207.08051},
  primaryclass = {cs},
  publisher = {arXiv},
  doi = {10.48550/arXiv.2207.08051},
  urldate = {2025-08-21},
  archiveprefix = {arXiv}
}

@misc{liu_swin_2021,
  title = {Swin {{Transformer}}: {{Hierarchical Vision Transformer}} Using {{Shifted Windows}}},
  shorttitle = {Swin {{Transformer}}},
  author = {Liu, Ze and Lin, Yutong and Cao, Yue and Hu, Han and Wei, Yixuan and Zhang, Zheng and Lin, Stephen and Guo, Baining},
  year = {2021},
  month = aug,
  number = {arXiv:2103.14030},
  eprint = {2103.14030},
  primaryclass = {cs},
  publisher = {arXiv},
  doi = {10.48550/arXiv.2103.14030},
  urldate = {2025-08-21},
  archiveprefix = {arXiv}
}

@article{stephens_cloudsat_2002,
  title = {{{THE CLOUDSAT MISSION AND THE A-TRAIN}}: {{A New Dimension}} of {{Space-Based Observations}} of {{Clouds}} and {{Precipitation}}},
  shorttitle = {{{THE CLOUDSAT MISSION AND THE A-TRAIN}}},
  author = {Stephens, Graeme L. and Vane, Deborah G. and Boain, Ronald J. and Mace, Gerald G. and Sassen, Kenneth and Wang, Zhien and Illingworth, Anthony J. and O'connor, Ewan J. and Rossow, William B. and Durden, Stephen L. and Miller, Steven D. and Austin, Richard T. and Benedetti, Angela and Mitrescu, Cristian},
  year = {2002},
  month = dec,
  journal = {Bulletin of the American Meteorological Society},
  volume = {83},
  number = {12},
  pages = {1771--1790},
  publisher = {American Meteorological Society},
  issn = {0003-0007, 1520-0477},
  doi = {10.1175/BAMS-83-12-1771},
  urldate = {2025-08-21},
  chapter = {Bulletin of the American Meteorological Society},
  langid = {english}
}

@article{winker_calipso_2010,
  title = {The {{CALIPSO Mission}}: {{A Global 3D View}} of {{Aerosols}} and {{Clouds}}},
  shorttitle = {The {{CALIPSO Mission}}},
  author = {Winker, D. M. and Pelon, J. and Coakley, J. A. and Ackerman, S. A. and Charlson, R. J. and Colarco, P. R. and Flamant, P. and Fu, Q. and Hoff, R. M. and Kittaka, C. and Kubar, T. L. and Treut, H. Le and Mccormick, M. P. and M{\'e}gie, G. and Poole, L. and Powell, K. and Trepte, C. and Vaughan, M. A. and Wielicki, B. A.},
  year = {2010},
  month = sep,
  journal = {Bulletin of the American Meteorological Society},
  volume = {91},
  number = {9},
  pages = {1211--1230},
  publisher = {American Meteorological Society},
  issn = {0003-0007, 1520-0477},
  doi = {10.1175/2010BAMS3009.1},
  urldate = {2025-08-21},
  chapter = {Bulletin of the American Meteorological Society},
  langid = {english}
}

@article{aminou_msg_2002,
  title = {{{MSG}}'s {{SEVIRI Instrument}}},
  author = {Aminou, D. M. A.},
  year = {2002},
  month = aug,
  journal = {ESA bulletin},
  volume = {111},
  pages = {15--17},
  urldate = {2024-05-01},
  langid = {english}
}

@article{schmit_closer_2016,
  title = {A {{Closer Look}} at the {{ABI}} on the {{GOES-R Series}}},
  author = {Schmit, Timothy J. and Griffith, Paul and Gunshor, Mathew M. and Daniels, Jaime M. and Goodman, Steven J. and Lebair, William J.},
  year = {2016},
  month = jul,
  journal = {Bulletin of the American Meteorological Society},
  volume = {98},
  number = {4},
  pages = {681--698},
  issn = {0003-0007},
  doi = {10.1175/BAMS-D-15-00230.1},
  urldate = {2020-02-13}
}

@article{bessho_introduction_2016,
  title = {An {{Introduction}} to {{Himawari-8}}/9--- {{Japan}}'s {{New-Generation Geostationary Meteorological Satellites}}},
  author = {Bessho, Kotaro and Date, Kenji and Hayashi, Masahiro and Ikeda, Akio and Imai, Takahito and Inoue, Hidekazu and Kumagai, Yukihiro and Miyakawa, Takuya and Murata, Hidehiko and Ohno, Tomoo and Okuyama, Arata and Oyama, Ryo and Sasaki, Yukio and Shimazu, Yoshio and Shimoji, Kazuki and Sumida, Yasuhiko and Suzuki, Masuo and Taniguchi, Hidetaka and Tsuchiyama, Hiroaki and Uesawa, Daisaku and Yokota, Hironobu and Yoshida, Ryo},
  year = {2016},
  journal = {Journal of the Meteorological Society of Japan. Ser. II},
  volume = {94},
  number = {2},
  pages = {151--183},
  doi = {10.2151/jmsj.2016-009}
}

@article{stephens_cloudsat_2018,
  title = {{{CloudSat}} and {{CALIPSO}} within the {{A-Train}}: {{Ten Years}} of {{Actively Observing}} the {{Earth System}}},
  shorttitle = {{{CloudSat}} and {{CALIPSO}} within the {{A-Train}}},
  author = {Stephens, Graeme and Winker, David and Pelon, Jacques and Trepte, Charles and Vane, Deborah and Yuhas, Cheryl and L'Ecuyer, Tristan and Lebsock, Matthew},
  year = {2018},
  month = mar,
  journal = {Bulletin of the American Meteorological Society},
  volume = {99},
  number = {3},
  pages = {569--581},
  publisher = {American Meteorological Society},
  issn = {0003-0007, 1520-0477},
  doi = {10.1175/BAMS-D-16-0324.1},
  urldate = {2023-05-24},
  chapter = {Bulletin of the American Meteorological Society},
  langid = {english}
}

@article{deng_cloudsat_2015,
  title = {{{CloudSat 2C-ICE}} Product Update with a New {{Ze}} Parameterization in Lidar-Only Region},
  author = {Deng, Min and Mace, {\relax Gerald}. G. and Wang, Zhien and Berry, Elizabeth},
  year = {2015},
  journal = {Journal of Geophysical Research: Atmospheres},
  volume = {120},
  number = {23},
  pages = {12,198--12,208},
  issn = {2169-8996},
  doi = {10.1002/2015JD023600},
  urldate = {2025-08-21},
  copyright = {{\copyright}2015. The Authors.},
  langid = {english}
}

@article{marchand_hydrometeor_2008,
  title = {Hydrometeor {{Detection Using Cloudsat}}---{{An Earth-Orbiting}} 94-{{GHz Cloud Radar}}},
  author = {Marchand, Roger and Mace, Gerald G. and Ackerman, Thomas and Stephens, Graeme},
  year = {2008},
  month = apr,
  journal = {Journal of Atmospheric and Oceanic Technology},
  volume = {25},
  number = {4},
  pages = {519--533},
  publisher = {American Meteorological Society},
  issn = {0739-0572, 1520-0426},
  doi = {10.1175/2007JTECHA1006.1},
  urldate = {2024-10-02},
  chapter = {Journal of Atmospheric and Oceanic Technology},
  langid = {english}
}

@misc{dosovitskiy_image_2021,
  title = {An {{Image}} Is {{Worth}} 16x16 {{Words}}: {{Transformers}} for {{Image Recognition}} at {{Scale}}},
  shorttitle = {An {{Image}} Is {{Worth}} 16x16 {{Words}}},
  author = {Dosovitskiy, Alexey and Beyer, Lucas and Kolesnikov, Alexander and Weissenborn, Dirk and Zhai, Xiaohua and Unterthiner, Thomas and Dehghani, Mostafa and Minderer, Matthias and Heigold, Georg and Gelly, Sylvain and Uszkoreit, Jakob and Houlsby, Neil},
  year = {2021},
  month = jun,
  number = {arXiv:2010.11929},
  eprint = {2010.11929},
  primaryclass = {cs},
  publisher = {arXiv},
  doi = {10.48550/arXiv.2010.11929},
  urldate = {2025-08-20},
  archiveprefix = {arXiv}
}

@article{francis_sensor_2024,
  title = {Sensor {{Independent Cloud}} and {{Shadow Masking With Partial Labels}} and {{Multimodal Inputs}}},
  author = {Francis, Alistair},
  year = {2024},
  journal = {IEEE Transactions on Geoscience and Remote Sensing},
  volume = {62},
  pages = {1--18},
  issn = {1558-0644},
  doi = {10.1109/TGRS.2024.3391625},
  urldate = {2025-08-20}
}

@article{hendricks_summary_2019,
  title = {A Summary of Research Advances on Tropical Cyclone Intensity Change from 2014-2018},
  author = {Hendricks, Eric A. and Braun, Scott A. and Vigh, Jonathan L. and Courtney, Joseph B.},
  year = {2019},
  month = dec,
  journal = {Tropical Cyclone Research and Review},
  volume = {8},
  number = {4},
  pages = {219--225},
  issn = {2225-6032},
  doi = {10.1016/j.tcrr.2020.01.002},
  urldate = {2025-08-20}
}

@article{leroux_recent_2018,
  title = {Recent {{Advances}} in {{Research}} and {{Forecasting}} of {{Tropical Cyclone Track}}, {{Intensity}}, and {{Structure}} at {{Landfall}}},
  author = {Leroux, Marie-Dominique and Wood, Kimberly and Elsberry, Russell L. and Cayanan, Esperanza O. and Hendricks, Eric and Kucas, Matthew and Otto, Peter and Rogers, Robert and Sampson, Buck and Yu, Zifeng},
  year = {2018},
  month = may,
  journal = {Tropical Cyclone Research and Review},
  volume = {7},
  number = {2},
  pages = {85--105},
  issn = {2225-6032},
  doi = {10.6057/2018TCRR02.02},
  urldate = {2025-08-20}
}

@incollection{sippel_tropical_2015,
  title = {{{TROPICAL CYCLONES AND HURRICANES}} {\textbar} {{Hurricane Predictability}}},
  booktitle = {Encyclopedia of {{Atmospheric Sciences}} ({{Second Edition}})},
  author = {Sippel, J. A.},
  editor = {North, Gerald R. and Pyle, John and Zhang, Fuqing},
  year = {2015},
  month = jan,
  pages = {30--34},
  publisher = {Academic Press},
  address = {Oxford},
  doi = {10.1016/B978-0-12-382225-3.00497-7},
  urldate = {2025-08-20},
  isbn = {978-0-12-382225-3}
}

@article{ruppert_critical_2020,
  title = {The Critical Role of Cloud--Infrared Radiation Feedback in Tropical Cyclone Development},
  author = {Ruppert, James H. and Wing, Allison A. and Tang, Xiaodong and Duran, Erika L.},
  year = {2020},
  month = nov,
  journal = {Proceedings of the National Academy of Sciences},
  volume = {117},
  number = {45},
  pages = {27884--27892},
  publisher = {Proceedings of the National Academy of Sciences},
  doi = {10.1073/pnas.2013584117},
  urldate = {2025-07-07}
}

@article{diakogiannis_resuneta_2020,
  title = {{{ResUNet-a}}: A Deep Learning Framework for Semantic Segmentation of Remotely Sensed Data},
  shorttitle = {{{ResUNet-a}}},
  author = {Diakogiannis, Foivos I. and Waldner, Fran{\c c}ois and Caccetta, Peter and Wu, Chen},
  year = {2020},
  month = apr,
  journal = {ISPRS Journal of Photogrammetry and Remote Sensing},
  volume = {162},
  eprint = {1904.00592},
  primaryclass = {cs},
  pages = {94--114},
  issn = {09242716},
  doi = {10.1016/j.isprsjprs.2020.01.013},
  urldate = {2025-08-20},
  archiveprefix = {arXiv}
}

@misc{ronneberger_unet_2015,
  title = {U-{{Net}}: {{Convolutional Networks}} for {{Biomedical Image Segmentation}}},
  shorttitle = {U-{{Net}}},
  author = {Ronneberger, Olaf and Fischer, Philipp and Brox, Thomas},
  year = {2015},
  month = may,
  number = {arXiv:1505.04597},
  eprint = {1505.04597},
  publisher = {arXiv},
  doi = {10.48550/arXiv.1505.04597},
  urldate = {2024-11-21},
  archiveprefix = {arXiv}
}

@misc{jeggle_icecloudnet_2024a,
  title = {{{IceCloudNet}}: {{3D}} Reconstruction of Cloud Ice from {{Meteosat SEVIRI}}},
  shorttitle = {{{IceCloudNet}}},
  author = {Jeggle, Kai and Czerkawski, Mikolaj and Serva, Federico and Saux, Bertrand Le and Neubauer, David and Lohmann, Ulrike},
  year = {2024},
  month = oct,
  number = {arXiv:2410.04135},
  eprint = {2410.04135},
  publisher = {arXiv},
  doi = {10.48550/arXiv.2410.04135},
  urldate = {2024-10-22},
  archiveprefix = {arXiv}
}

@article{bruning_artificial_2024,
  title = {Artificial Intelligence ({{AI}})-Derived {{3D}} Cloud Tomography from Geostationary {{2D}} Satellite Data},
  author = {Br{\"u}ning, Sarah and Niebler, Stefan and Tost, Holger},
  year = {2024},
  month = feb,
  journal = {Atmospheric Measurement Techniques},
  volume = {17},
  number = {3},
  pages = {961--978},
  publisher = {Copernicus GmbH},
  issn = {1867-1381},
  doi = {10.5194/amt-17-961-2024},
  urldate = {2024-10-22},
  langid = {english}
}

@misc{noaa_us_2023,
    title = {U.{S}. {Billion}-{Dollar} {Weather} \& {Climate} {Disasters} 1980-2022},
    url = {https://www.ncei.noaa.gov/access/billions/events.pdf},
    urldate = {2023-03-14},
    author = {{NOAA}},
    year = {2023},
}

@article{elsberry_new_2025,
    title = {New {Challenges} for {Tropical} {Cyclone} {Track} and {Intensity} {Forecasting} in {Unfavorable} {External} {Environment} in {Western} {North} {Pacific}. {Part} {I}. {Formations} {South} of 20° {N}},
    volume = {16},
    copyright = {https://creativecommons.org/licenses/by/4.0/},
    issn = {2073-4433},
    url = {https://www.mdpi.com/2073-4433/16/2/226},
    doi = {10.3390/atmos16020226},
    language = {en},
    number = {2},
    urldate = {2025-08-12},
    journal = {Atmosphere},
    author = {Elsberry, Russell L. and Tsai, Hsiao-Chung and Huang, Wen-Hsin and Marchok, Timothy P.},
    month = feb,
    year = {2025},
    pages = {226},
}

@article{lee_satellite-based_2024,
    title = {Satellite-{Based} {Estimation} of the {Role} of {Cloud}-{Radiative} {Interaction} in {Accelerating} {Tropical} {Cyclone} {Development}},
    volume = {81},
    copyright = {http://www.ametsoc.org/PUBSReuseLicenses},
    issn = {0022-4928, 1520-0469},
    url = {https://journals.ametsoc.org/view/journals/atsc/81/6/JAS-D-23-0142.1.xml},
    doi = {10.1175/JAS-D-23-0142.1},
    number = {6},
    urldate = {2025-08-12},
    journal = {Journal of the Atmospheric Sciences},
    author = {Lee, Tsung-Yung and Wing, Allison A.},
    month = jun,
    year = {2024},
    pages = {959--982},
}

@article{wing_acceleration_2022,
    title = {Acceleration of {Tropical} {Cyclone} {Development} by {Cloud}-{Radiative} {Feedbacks}},
    volume = {79},
    issn = {0022-4928, 1520-0469},
    url = {https://journals.ametsoc.org/view/journals/atsc/79/9/JAS-D-21-0227.1.xml},
    doi = {10.1175/JAS-D-21-0227.1},
    number = {9},
    urldate = {2025-08-12},
    journal = {Journal of the Atmospheric Sciences},
    author = {Wing, Allison A.},
    month = sep,
    year = {2022},
    pages = {2285--2305},
}

@misc{girtsou_3d_2025,
    title = {{3D} {Cloud} reconstruction through geospatially-aware {Masked} {Autoencoders}},
    url = {http://arxiv.org/abs/2501.02035},
    doi = {10.48550/arXiv.2501.02035},
    urldate = {2025-08-12},
    publisher = {arXiv},
    author = {Girtsou, Stella and Salas-Porras, Emiliano Diaz and Freischem, Lilli and Massant, Joppe and Bintsi, Kyriaki-Margarita and Castiglione, Guiseppe and Jones, William and Eisinger, Michael and Johnson, Emmanuel and Jungbluth, Anna},
    month = jan,
    year = {2025},
    note = {arXiv:2501.02035 [cs]},
}

@inproceedings{Kingma2015Adam:Optimization,
    title = {{Adam: A method for stochastic optimization}},
    year = {2015},
    booktitle = {3rd International Conference on Learning Representations, ICLR 2015 - Conference Track Proceedings},
    author = {Kingma, Diederik P. and Ba, Jimmy Lei},
    arxivId = {1412.6980}
}

@article{Rumelhart1986LearningErrors,
    title = {{Learning representations by back-propagating errors}},
    year = {1986},
    journal = {Nature},
    author = {Rumelhart, David E. and Hinton, Geoffrey E. and Williams, Ronald J.},
    number = {6088},
    pages = {533--536},
    volume = {323},
    publisher = {Nature Publishing Group},
    doi = {10.1038/323533a0},
    issn = {00280836}
}

@article{xu2023swin,
  title={Swin MAE: Masked autoencoders for small datasets},
  author={Xu, Zi’an and Dai, Yin and Liu, Fayu and Chen, Weibing and Liu, Yue and Shi, Lifu and Liu, Sheng and Zhou, Yuhang},
  journal={Computers in biology and medicine},
  volume={161},
  pages={107037},
  year={2023},
  publisher={Elsevier}
}

@article{Barker2011,
  title = {A 3D cloud‐construction algorithm for the EarthCARE satellite mission},
  volume = {137},
  ISSN = {1477-870X},
  url = {http://dx.doi.org/10.1002/qj.824},
  DOI = {10.1002/qj.824},
  number = {657},
  journal = {Quarterly Journal of the Royal Meteorological Society},
  publisher = {Wiley},
  author = {Barker,  H. W. and Jerg,  M. P. and Wehr,  T. and Kato,  S. and Donovan,  D. P. and Hogan,  R. J.},
  year = {2011},
  month = apr,
  pages = {1042–1058}
}

@article{Horvth2001,
  title = {Simultaneous retrieval of cloud motion and height from polar‐orbiter multiangle measurements},
  volume = {28},
  ISSN = {1944-8007},
  url = {http://dx.doi.org/10.1029/2001gl012951},
  DOI = {10.1029/2001gl012951},
  number = {15},
  journal = {Geophysical Research Letters},
  publisher = {American Geophysical Union (AGU)},
  author = {Horváth,  Ákos and Davies,  Roger},
  year = {2001},
  month = aug,
  pages = {2915–2918}
}

@article{Volkmer2024,
  title = {Consideration of the cloud motion for aircraft-based stereographically derived cloud geometry and cloud top heights},
  volume = {17},
  ISSN = {1867-8548},
  url = {http://dx.doi.org/10.5194/amt-17-6807-2024},
  DOI = {10.5194/amt-17-6807-2024},
  number = {23},
  journal = {Atmospheric Measurement Techniques},
  publisher = {Copernicus GmbH},
  author = {Volkmer,  Lea and K\"{o}lling,  Tobias and Zinner,  Tobias and Mayer,  Bernhard},
  year = {2024},
  month = dec,
  pages = {6807–6817}
}

@misc{orcestra,
    title = {ORCESTRA - Organized Convection and EarthCARE Studies over the Tropical Atlantic},
    author = {Klocke, Daniel and  Dengler, Marcus and Carlsen, Tim  and David, Robert Oscar and Baars, Holger and Skupin, Annett and Bony, Sandrine and Delanoë, Julien and  Gross, Silke and Stevens, Bjorn and Windmiller, Julia and Wing, Allison and Vogel, Raphaela and George, Geet},
    howpublished = {\url{https://orcestra-campaign.org/}},
}

@article{ARM,
   author    =  "G. M. Stokes and S. E. Schwartz",
   title     =  "The Atmospheric Radiation Measurement (ARM) Program: Programmatic background and design of the cloud and radiation test bed",
   year      =  "1994",
   journal   =  "Bull. Amer. Meteor. Soc.",
   volume    =  "75",
   pages     =  "1201--1222"
}

@Article{earthCARE,
AUTHOR = {Wehr, T. and Kubota, T. and Tzeremes, G. and Wallace, K. and Nakatsuka, H. and Ohno, Y. and Koopman, R. and Rusli, S. and Kikuchi, M. and Eisinger, M. and Tanaka, T. and Taga, M. and Deghaye, P. and Tomita, E. and Bernaerts, D.},
TITLE = {The EarthCARE mission -- science and system overview},
JOURNAL = {Atmospheric Measurement Techniques},
VOLUME = {16},
YEAR = {2023},
NUMBER = {15},
PAGES = {3581--3608},
URL = {https://amt.copernicus.org/articles/16/3581/2023/},
DOI = {10.5194/amt-16-3581-2023}
}

\clearpage

\section*{Appendix}
\label{Appendix}

\subsection*{Dataset Details}

\begin{table}[htb]
\caption{Details of our pre-training dataset. We processed 50,000 patches of $1024 \times 1024$ that are randomly cropped to $256 \times 256$ pixels in each training step.}
\label{appendix-table-pre-training-dataset}
\begin{tabular}{|l|l|l|l|l|l|l|}
\hline
         & \textbf{Coverage} & \textbf{Product} & \textbf{Resolution} & \textbf{\# Channels} & \textbf{\# Files} & \textbf{Size}   \\ \hline
\textbf{GOES}     & 2018 - 2024       & MCMIP   & 2 km @ SSP & 16          & 50,000   & 793 GB \\ \hline
\textbf{MSG}      & 2004 - 2025       & L1b     & 3 km @ SSP & 11          & 50,000   & 646 GB \\ \hline
\textbf{HIMAWARI} & 2015 - 2022       & L1b     & 2 km @ SSP & 16          & 50,000   & 930 GB \\ \hline
\end{tabular}
\end{table}

\begin{table}[htb]
\caption{Details of our clouds dataset. We aligned the geostationary satellite imagery and corresponding CloudSat overpasses in space and time, and save crops of $256 \times 256$ pixels.}
\label{appendix-table-finetuning-dataset}
\begin{tabular}{|l|l|l|l|l|l|}
\hline
                           & \textbf{Coverage} & \textbf{Resolution} & \textbf{\# Channels} & \textbf{\# Files} & \textbf{Size} \\ \hline
\textbf{GOES-CloudSat}     & 2018 - 2020                & 2 km @ SSP          & 16                   & 31,046            & 72 GB         \\ \hline
\textbf{MSG-CloudSat}      & 2006 - 2020                & 3 km @ SSP          & 11                   & 181,653           & 620 GB        \\ \hline
\textbf{HIMAWARI-CloudSat} & 2015 - 2020                & 2 km @ SSP          & 16                   & 57,373            & 181 GB        \\ \hline
\end{tabular}
\end{table}

\begin{table}[htb]
\caption{We aligned the geostationary satellite imagery of tropical cyclones and corresponding CloudSat overpassed in space and time, based on TC track information provided by the International Best Track Archive for Climate Stewardship (IBTrACS). We crop $256 \times 256$ pixels around each overpass. We consider all CloudSat overpasses within 256 km of the TC center. We focus on storms in the North Atlantic and Eastern Pacific for GOES, and the Western and Southern Pacific for HIMAWARI. Note that there are no cyclones in the MSG field-of-view. Our TC dataset is reserved exclusively for evaluation, providing a rigorous testbed for assessing model performance under the most intense storm conditions.}
\label{appendix-table-TC-dataset}
\begin{tabular}{|l|l|l|l|l|l|}
\hline
                           & \textbf{Coverage} & \textbf{Resolution} & \textbf{\# Channels} & \textbf{\# Files} & \textbf{Size} \\ \hline
\textbf{GOES-CloudSat}     & 2018 - 2020                & 2 km @ SSP          & 16                   & 185               & 341 MB        \\ \hline
\textbf{MSG-CloudSat}      & N/A                        & 3 km @ SSP          & 11                   & N/A               & N/A           \\ \hline
\textbf{HIMAWARI-CloudSat} & 2015 - 2020                & 2 km @ SSP          & 16                   & 518               & 1.4 GB        \\ \hline
\end{tabular}
\end{table}

\subsection*{Training Details}\label{appendix-training-details}

We conducted our experiments on a single NVIDIA V100 GPU via Google Cloud, with batch size of 32 during pre-training and 8-16 during fine-tuning. We used the Adam optimizer \cite{Kingma2015Adam:Optimization} with a learning rate of 0.00015, using backpropagation \cite{Rumelhart1986LearningErrors}.
Training was optimized via the Mean Squared Error (MSE) loss, while additional metrics like the Peak Signal-to-Noise Ratio (PSNR) and the Structural Similarity Index Measure (SSIM) were monitored. Checkpointing was used to save models with the lowest validation loss. Pre-training and fine-tuning ran for 50 and 100 epochs respectively. Regarding training times, the pre-training took around 6 hours, while fine-tuning ran for 5 hours - 3 days depending on the complexity of the model.
Our U-Net baseline contains 1.9M trainable parameters, while our SWinSatMAE model contains 34.2M trainable parameters. Increasing the complexity of the U-Net to match or exceed the parameters of our model did not improve prediction results. 
Our U-Net was trained without dropout, $10^{-5}$ weight decay, and 4 up- and down-blocks of residual convolutions. Our SWinSatMAE model was pre-trained with a token size of $2 \times 2$ pixels, a masking window of $4 \times 4$ pixels, an attention window of $32 \times 32$ pixels, and 50\% masking. During fine-tuning, we replaced the pre-training decoder with a custom SwinConv decoder that first reverses the operations of the Swin encoder, and then uses repeated blocks of ConvTranspose2D, ResidualConv, and ReLU activations to reach the final desired output of $256 \times 256$ pixels and 96 height levels. We then predict the $n$ target variables and the final 80 height levels using $n$ prediction heads, each made up of sequentially applied residual convolution, ReLu, convolution operationns.
We encode the latitude/longitude coordinates, time of measurement (fraction of the day and fraction of the year), satellite viewing angle (zenith and azimuth), and solar angle (zenith and azimuth) together with the positional encoding to make our model geospatially aware. 
We split our dataset by time, and allocate days 2-22 each month to training, 24 - 26 to validation, and 28 - 31 to testing. We purposely leave a gap of 1 day to avoid data leakage. During fine-tuning on our clouds dataset, we filter our examples that contain less than 25\% of cloudy columns.


\subsection*{Image Reconstructions During Pretraining}

\begin{figure}[H]
    \centering
    \includegraphics[width=0.7\linewidth]{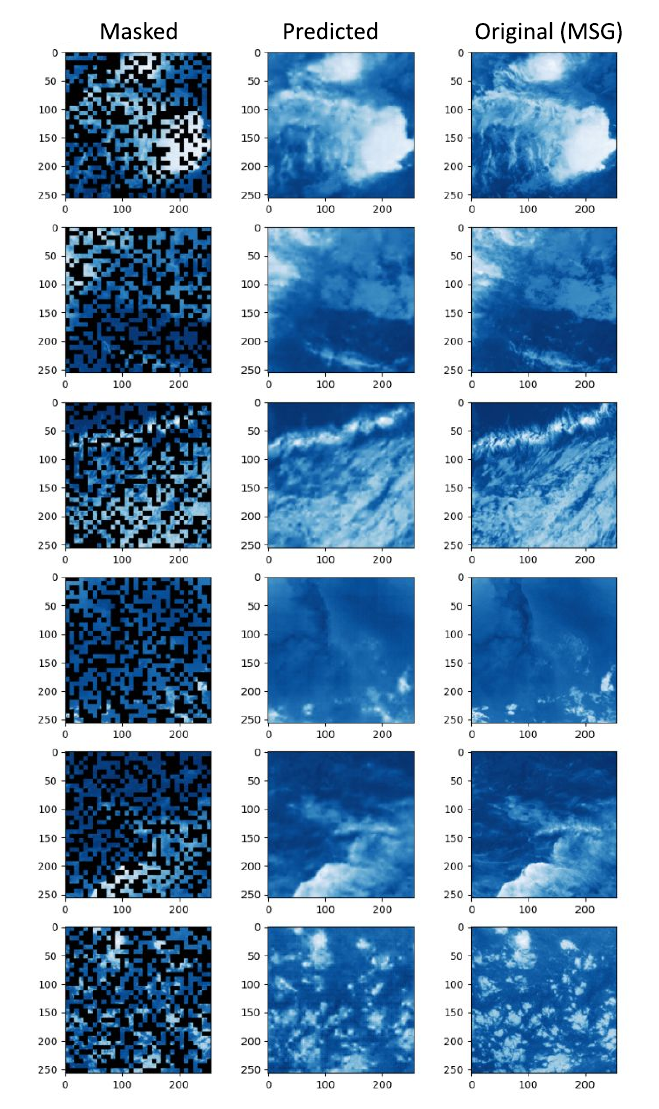}
    \caption{Comparison of masked, predicted, and original images (MSG example) during pre-training of our SWinSatMAE model. We pretrained our model for 50 epochs.}
    \label{appendix-figure-pre-training}
\end{figure}

\clearpage

\subsection*{U-Net: Single Satellite vs. Multi-Satellite Input}

\begin{table}[thb]
\caption{Comparison of a U-Net model, trained to predict Z using (a/c) a single satellite or (b/d) all three satellites as model input. Sub-tables (a) and (b) show the performance on our clouds test set, while (c) and (d) show the performance on our entire cyclone dataset, which was not seen during training. 
When training on each satellite individually, we use all spectral channels, i.e. 16 spectral channels for GOES and HIMAWARI, and 11 spectral channels for MSG. To combine the satellites into a unified dataset, we match the closest 11 spectral channels from each satellite and ignore resolution differences.
We report the root-mean-squared error (RMSE; in dBZ), structural similarity index measure (SSIM; unitless) and Dice coefficient (unitless). All models were trained for 100 epochs, and the best validation loss checkpoint was chosen for inference.
Despite differences in spectral characteristics and resolution of the different satellites, naive matching of satellites and spectral channels does not degrade model performance.}
\label{appendix-table-singlesat-multisat}

\vspace{1em}
\begin{flushleft}
\includegraphics[height=2em]{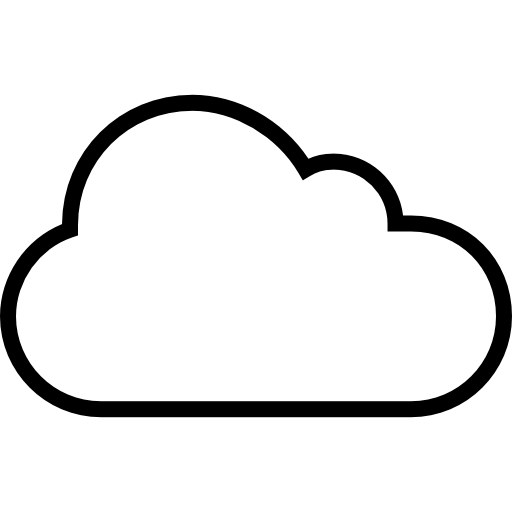} \\
\begin{tabular}{|L{0.2cm}L{2.15cm}|C{2.3cm}|C{2.3cm}|C{2.3cm}|}
\hline
\textbf{(a)} &
\textbf{Model Input} & \multicolumn{3}{c|}{\textbf{Single Satellite}} \\ \hline
& \textbf{Satellite} & GOES & MSG & HIMAWARI \\ \hline
& RMSE (dBZ) & 5.44 $\pm$ 2.61 & \textbf{4.81 $\pm$ 2.21} & 6.14 $\pm$ 2.99 \\ 
& SSIM & 0.75 $\pm$ 0.13 & \textbf{0.78 $\pm$ 0.11} & 0.71 $\pm$ 0.14 \\ 
& DICE & 0.75 $\pm$ 0.13 & \textbf{0.88 $\pm$ 0.15} & 0.84 $\pm$ 0.16 \\ \hline
\end{tabular}
\end{flushleft}

\vspace{-0.5em}
\begin{flushleft}
\begin{tabular}{|L{0.2cm}L{2.15cm}|C{2.3cm}|C{2.3cm}|C{2.3cm}|C{2.2cm}|}
\hline
\textbf{(b)} & \textbf{Model Input} & \multicolumn{4}{c|}{\textbf{Multi-Satellite}} \\ \hline
& \textbf{Satellite} & GOES & MSG & HIMAWARI & Combined \\ \hline
& RMSE (dBZ) & \textbf{5.22$\pm$2.65} & 4.95$\pm$2.17 & \textbf{6.03$\pm$3.05} & 5.40$\pm$2.69 \\ 
& SSIM & \textbf{0.76$\pm$0.13} & 0.77$\pm$0.11 & \textbf{0.72$\pm$0.15} & 0.75$\pm$0.13 \\ 
& DICE & \textbf{0.84$\pm$0.18} & 0.87$\pm$0.17 & \textbf{0.83$\pm$0.17} & 0.85$\pm$0.17 \\ \hline
\end{tabular}
\hfill
\end{flushleft}

\vspace{0.5em}
\begin{flushleft}
\includegraphics[height=2em]{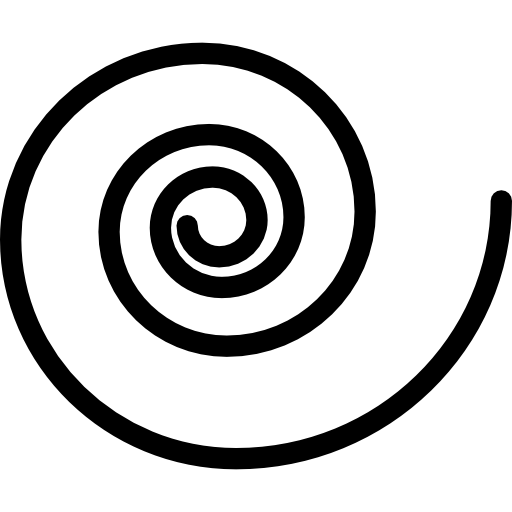} \\
\begin{tabular}{|L{0.2cm}L{2.15cm}|C{2.3cm}|C{2.3cm}|C{2.3cm}|}
\hline
\textbf{(c)} & \textbf{Model Input} & \multicolumn{3}{c|}{\textbf{Single Satellite}} \\ \hline
& \textbf{Satellite} & GOES & MSG & HIMAWARI \\ \hline
& RMSE (dBZ) & \textbf{7.65$\pm$4.42} & N/A & 10.82$\pm$5.01 \\ 
& SSIM & \textbf{0.65$\pm$0.21} & N/A & 0.47$\pm$0.22 \\ 
& DICE & \textbf{0.82$\pm$0.2}  & N/A & 0.84$\pm$0.15 \\ \hline
\end{tabular}
\end{flushleft}

\vspace{-0.5em}
\begin{flushleft}
\begin{tabular}{|L{0.2cm}L{2.15cm}|C{2.3cm}|C{2.3cm}|C{2.3cm}|C{2.2cm}|}
\hline
\textbf{(d)} & \textbf{Model Input} & \multicolumn{4}{c|}{\textbf{Multi-Satellite}} \\ \hline
& \textbf{Satellite} & GOES & MSG & HIMAWARI & Combined \\ \hline
& RMSE (dBZ) & 8.24$\pm$4.95 & N/A & \textbf{9.41$\pm$4.33} & 8.89$\pm$4.62 \\ 
& SSIM & 0.62$\pm$0.22 & N/A & \textbf{0.5$\pm$0.21 } & 0.55$\pm$0.22 \\ 
& DICE & 0.77$\pm$0.28 & N/A & \textbf{0.84$\pm$0.17} & 0.81$\pm$0.23 \\ \hline
\end{tabular}
\end{flushleft}
\end{table}

\begin{figure}
    \centering
    \includegraphics[width=0.7\linewidth]{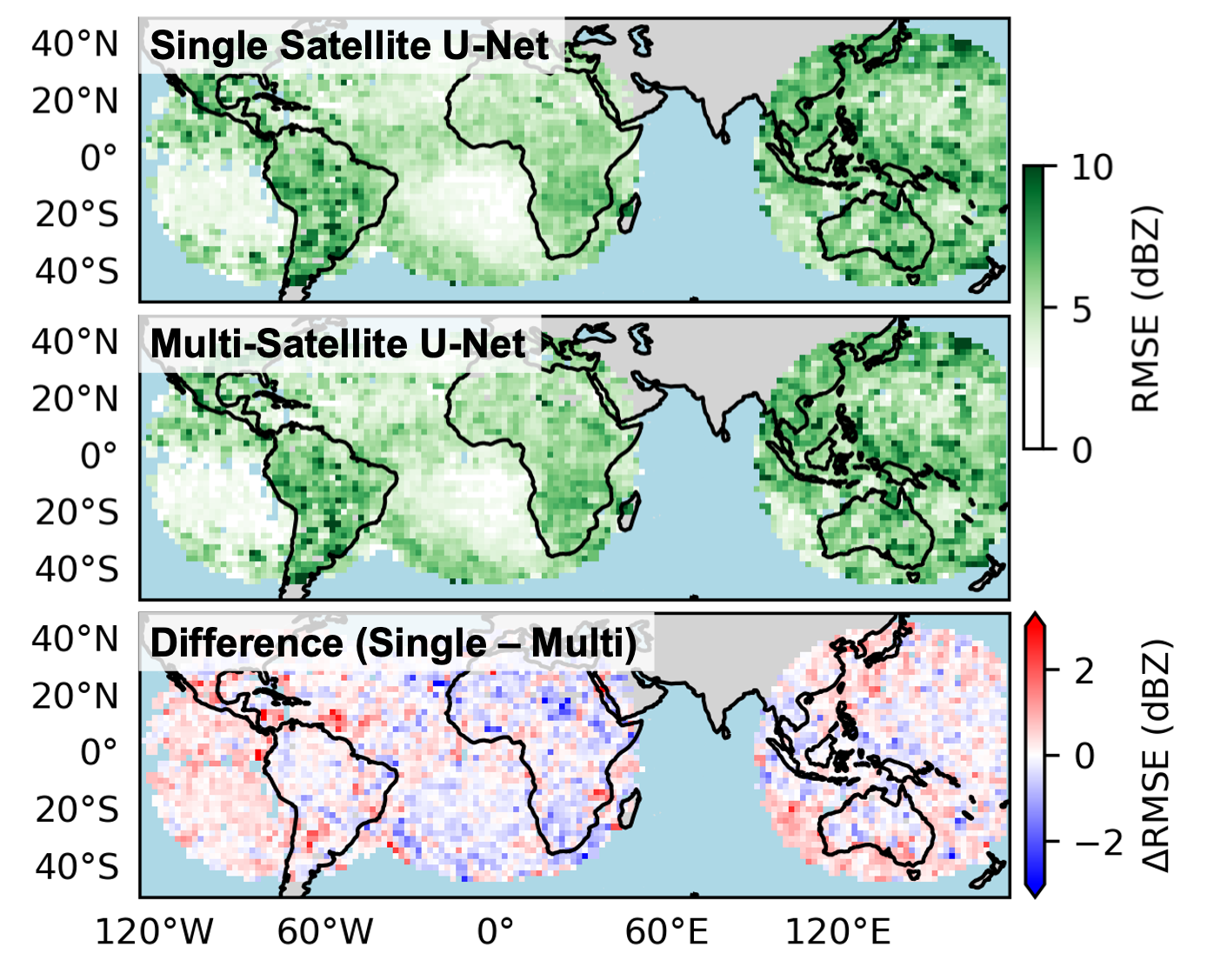}
    \caption{Comparison of prediction performance across the globe between U-Net models trained on our three satellites, MSG, GOES and Himawari individually (top), and a U-Net model trained on all three satellites together (Multi-Satellite U-Net, middle). }
    \label{appendix-map-singlesat-multisat}
\end{figure}

\clearpage

\subsection*{U-Net: Single Variable vs. Multi-Variable Target}

\begin{table}[htb]
\caption{Comparison of a U-Net model, trained to predict either one variable (a/c) or all three variables jointly (b/d). Sub-tables (a) and (b) show the performance on our clouds test set, while (c) and (d) show the performance on our entire cyclone dataset, which was not seen during training. All models are trained on the closest matched spectral channels for GOES, MSG, and HIMAWARI. We report the root-mean-squared error (RMSE; in variable units), structural similarity index measure (SSIM; unitless), and Dice coefficient (unitless). All models were trained for 100 epochs, and the best validation loss checkpoint was chosen for inference.}
\label{appendix-table-singlevar-multivar}

\vspace{1em}
\begin{flushleft}
\includegraphics[height=2em]{figures/icon_cloud.png} \\
\begin{tabular}{|L{0.2cm}L{2.1cm}|C{3.35cm}|C{3.4cm}|C{2.95cm}|}
\hline
\textbf{(a)} & \textbf{Model Target} & \multicolumn{3}{c|}{\textbf{Single Variable}} \\ \hline
& \textbf{Variable} & Z (dBZ) & Ice Water Content (g/m\textsuperscript{3}) & Effective Radius (\textmu m) \\ \hline
& RMSE (units) & 5.40 $\pm$ 2.69 & 0.07 $\pm$ 0.08 & 10.99 $\pm$ 5.20 \\ 
& SSIM & 0.75 $\pm$ 0.13 & \textbf{0.91 $\pm$ 0.07} & 0.52 $\pm$ 0.11 \\ 
& DICE & \textbf{0.85 $\pm$ 0.17} & \textbf{0.30 $\pm$ 0.09} & \textbf{0.03 $\pm$ 0.05} \\ \hline
\end{tabular}
\end{flushleft}

\vspace{-0.5em}
\begin{flushleft}
\begin{tabular}{|L{0.2cm}L{2.1cm}|C{3.35cm}|C{3.4cm}|C{2.95cm}|}
\hline
\textbf{(b)} & \textbf{Model Target} & \multicolumn{3}{c|}{\textbf{Multi-Variable}} \\ \hline
& \textbf{Variable} & Z (dBZ) & Ice Water Content (g/m\textsuperscript{3}) & Effective Radius (\textmu m) \\ \hline
& RMSE (units) & \textbf{5.10 $\pm$ 2.46} & \textbf{0.06 $\pm$ 0.08} & \textbf{10.30 $\pm$ 5.21} \\ 
& SSIM & \textbf{0.77 $\pm$ 0.13} & \textbf{0.91 $\pm$ 0.07} & \textbf{0.55 $\pm$ 0.11} \\ 
& DICE & \textbf{0.85 $\pm$ 0.17} & \textbf{0.30 $\pm$ 0.09} & \textbf{0.03 $\pm$ 0.04} \\ \hline
\end{tabular}
\end{flushleft}

\vspace{0.5em}
\begin{flushleft}
\includegraphics[height=2em]{figures/icon_hurricane.png} \\
\begin{tabular}{|L{0.2cm}L{2.1cm}|C{3.35cm}|C{3.4cm}|C{2.95cm}|}
\hline
\textbf{(c)} & \textbf{Model Target} & \multicolumn{3}{c|}{\textbf{Single Variable}} \\ \hline
& \textbf{Variable} & Z (dBZ) & Ice Water Content (g/m\textsuperscript{3}) & Effective Radius (\textmu m) \\ \hline
& RMSE (units) & 9.03 $\pm$ 4.95 & \textbf{0.16 $\pm$ 0.16} & 14.80 $\pm$ 7.99 \\ 
& SSIM & \textbf{0.57 $\pm$ 0.24} & \textbf{0.84 $\pm$ 0.12} & 0.48 $\pm$ 0.13 \\ 
& DICE & \textbf{0.82 $\pm$ 0.20} & \textbf{0.31 $\pm$ 0.07} & \textbf{0.03 $\pm$ 0.03} \\ \hline
\end{tabular}
\end{flushleft}

\vspace{-0.5em}
\begin{flushleft}
\begin{tabular}{|L{0.2cm}L{2.1cm}|C{3.35cm}|C{3.4cm}|C{2.95cm}|}
\hline
\textbf{(d)} & \textbf{Model Target} & \multicolumn{3}{c|}{\textbf{Multi-Variable}} \\ \hline
& \textbf{Variable} & Z (dBZ) & Ice Water Content (g/m\textsuperscript{3}) & Effective Radius (\textmu m) \\ \hline
& RMSE (units) & \textbf{8.55 $\pm$ 4.39} & 0.17 $\pm$ 0.16 & \textbf{14.66 $\pm$ 7.72} \\ 
& SSIM & \textbf{0.57 $\pm$ 0.24} & 0.83 $\pm$ 0.12 & \textbf{0.49 $\pm$ 0.14} \\ 
& DICE & 0.80 $\pm$ 0.19 & \textbf{0.31 $\pm$ 0.07} & \textbf{0.03 $\pm$ 0.04} \\ \hline
\end{tabular}
\end{flushleft}
\end{table}

\clearpage

\subsection*{Advanced Model Comparison}

\begin{table}[ht]
\caption{\textbf{Clear \& Cloudy}: Comparison of a U-Net model, a SWinMAE without encodings, our SWinSatMAE trained from scratch and including pre-training, across clear and cloudy pixels. We report the root-mean-squared error (RMSE; variable units) for (a) our clouds test set, and (b) for our entire cyclone dataset, which was not seen during training. All models are trained on the closest matched spectral channels for GOES, MSG, and HIMAWARI. All models were trained for 100 epochs, and the best validation loss checkpoint was chosen for inference.}
\label{appendix-table-comparison-all}
\begin{flushleft}
\includegraphics[height=2em]{figures/icon_cloud.png} \\
\begin{tabular}{|L{0.2cm}L{2.6cm}|C{3.0cm}|C{3.0cm}|C{3.05cm}|}
\hline
\textbf{(a)} & & \textbf{\begin{tabular}[c]{@{}c@{}}Z\\RMSE (dBZ)\end{tabular}} & 
\textbf{\begin{tabular}[c]{@{}c@{}}Ice Water Content\\RMSE (g/m\textsuperscript{3})\end{tabular}} & \textbf{\begin{tabular}[c]{@{}c@{}}Effective Radius\\RMSE (\textmu m)\end{tabular}} \\ \hline
& U-Net Baseline & 5.10 $\pm$ 2.46 & 0.063 $\pm$ 0.081 & 10.3 $\pm$ 5.21 \\
\hline
& SWinMAE & 4.21 $\pm$ 2.01 & 0.055 $\pm$ 0.071 & 7.93 $\pm$ 4.26 \\
\hline
& \begin{tabular}[c]{@{}l@{}}SWinSatMAE\\(no pre-training)\end{tabular} & 4.11 $\pm$ 1.98 & 0.055 $\pm$ 0.070 & 7.76 $\pm$ 4.21 \\
\hline
& \begin{tabular}[c]{@{}l@{}}SWinSatMAE\\(our model)\end{tabular} & \textbf{4.04 $\pm$ 1.93} & \textbf{0.053 $\pm$ 0.069} & \textbf{7.72 $\pm$ 3.99} \\ 
\hline
\end{tabular}
\end{flushleft}

\vspace{-0.5em}
\begin{flushleft}
\includegraphics[height=2em]{figures/icon_hurricane.png} \\
\begin{tabular}{|L{0.2cm}L{2.6cm}|C{3.0cm}|C{3.0cm}|C{3.05cm}|}
\hline
\textbf{(b)} & & \textbf{\begin{tabular}[c]{@{}c@{}}Z\\RMSE (dBZ)\end{tabular}} & 
\textbf{\begin{tabular}[c]{@{}c@{}}Ice Water Content\\RMSE (g/m\textsuperscript{3})\end{tabular}} & \textbf{\begin{tabular}[c]{@{}c@{}}Effective Radius\\RMSE (\textmu m)\end{tabular}} \\ \hline
& U-Net Baseline & 8.55 $\pm$ 4.39 & 0.166 $\pm$ 0.160 & 14.66 $\pm$ 7.72 \\
\hline
& SWinMAE & \textbf{7.01 $\pm$ 2.61} & \textbf{0.144 $\pm$ 0.134} & \textbf{11.53 $\pm$ 5.12} \\
\hline
& \begin{tabular}[c]{@{}l@{}}SWinSatMAE\\(no pre-training)\end{tabular} & 7.08 $\pm$ 2.76 & \textbf{0.144 $\pm$ 0.136} & 11.74 $\pm$ 5.46 \\
\hline
& \begin{tabular}[c]{@{}l@{}}SWinSatMAE\\(our model)\end{tabular} & 7.08 $\pm$ 2.62 & 0.147 $\pm$ 0.148 & 11.55 $\pm$ 5.03 \\ 
\hline
\end{tabular}
\end{flushleft}
\end{table}

\begin{table}[ht]
\caption{\textbf{Cloudy only}: Comparison of a U-Net model, a SWinMAE without encodings, our SWinSatMAE trained from scratch and including pre-training, across only cloudy pixels. We report the root-mean-squared error (RMSE; variable units) for (a) our clouds test set, and (b) for our entire cyclone dataset, which was not seen during training. All models are trained on the closest matched spectral channels for GOES, MSG, and HIMAWARI. All models were trained for 100 epochs, and the best validation loss checkpoint was chosen for inference.}
\label{appendix-table-comparison-cloudy}

\begin{flushleft}
\includegraphics[height=2em]{figures/icon_cloud.png} \\
\begin{tabular}{|L{0.2cm}L{2.6cm}|C{3.0cm}|C{3.0cm}|C{3.05cm}|}
\hline
\textbf{(a)} & & \textbf{\begin{tabular}[c]{@{}c@{}}Z\\RMSE (dBZ)\end{tabular}} & 
\textbf{\begin{tabular}[c]{@{}c@{}}Ice Water Content\\RMSE (g/m\textsuperscript{3})\end{tabular}} & \textbf{\begin{tabular}[c]{@{}c@{}}Effective Radius\\RMSE (\textmu m)\end{tabular}} \\ \hline
& U-Net Baseline & 9.50 $\pm$ 5.30 & 0.082 $\pm$ 0.164 & 14.79 $\pm$ 13.08 \\
\hline
& SWinMAE & 8.79 $\pm$ 5.11 & 0.075 $\pm$ 0.149 & 13.08 $\pm$ 11.47 \\
\hline
& \begin{tabular}[c]{@{}l@{}}SWinSatMAE\\(no pre-training)\end{tabular} & 8.57 $\pm$ 5.01 & 0.074 $\pm$ 0.149 & 12.91 $\pm$ 11.18 \\
\hline
& \begin{tabular}[c]{@{}l@{}}SWinSatMAE\\(our model)\end{tabular} & \textbf{8.46 $\pm$ 4.99} & \textbf{0.072 $\pm$ 0.145} & \textbf{12.40 $\pm$ 10.79} \\ 
\hline
\end{tabular}
\end{flushleft}

\vspace{-0.5em}
\begin{flushleft}
\includegraphics[height=2em]{figures/icon_hurricane.png} \\
\begin{tabular}{|L{0.2cm}L{2.6cm}|C{3.0cm}|C{3.0cm}|C{3.05cm}|}
\hline
\textbf{(c)} & & \textbf{\begin{tabular}[c]{@{}c@{}}Z\\RMSE (dBZ)\end{tabular}} & 
\textbf{\begin{tabular}[c]{@{}c@{}}Ice Water Content\\RMSE (g/m\textsuperscript{3})\end{tabular}} & \textbf{\begin{tabular}[c]{@{}c@{}}Effective Radius\\RMSE (\textmu m)\end{tabular}} \\ \hline
& U-Net Baseline & 12.75 $\pm$ 6.18 & 0.111 $\pm$ 0.179 & 16.52 $\pm$ 12.53 \\
\hline
& SWinMAE & \textbf{11.40 $\pm$ 5.58} & 0.104 $\pm$ 0.160 & 13.70 $\pm$ 10.70 \\
\hline
& \begin{tabular}[c]{@{}l@{}}SWinSatMAE\\(no pre-training)\end{tabular} & 11.47 $\pm$ 5.61 & 0.103 $\pm$ 0.159 & 14.27 $\pm$ 11.10 \\
\hline
& \begin{tabular}[c]{@{}l@{}}SWinSatMAE\\(our model)\end{tabular} & 11.50 $\pm$ 5.65 & \textbf{0.101 $\pm$ 0.160} & \textbf{13.21 $\pm$ 10.49} \\ 
\hline
\end{tabular}
\end{flushleft}
\end{table}

\clearpage

\begin{table}[htb]
\caption{Performance comparison across different cloud types. We report the root-mean-squared error (RMSE; variable units) for each cloud classification category in our clouds test set. All models were trained for 100 epochs, and the best validation loss checkpoint was chosen for inference.}
\label{appendix-table-cloud-types-comparison}
\vspace{1em}
\begin{flushleft}
\includegraphics[height=2em]{figures/icon_cloud.png} \\
\begin{tabular}{|L{2.4cm}|C{2.3cm}|C{2.3cm}|C{2.3cm}|C{2.3cm}|}
\hline
 & \multicolumn{4}{c|}{\textbf{Z RMSE (dBZ)}} \\ \hline
\textbf{Cloud Type} & \textbf{U-Net Baseline} & \textbf{SWinMAE} & \textbf{\begin{tabular}[c]{@{}c@{}}SWinSatMAE\\(no pre-training)\end{tabular}} & \textbf{\begin{tabular}[c]{@{}c@{}}SWinSatMAE\\(our model)\end{tabular}} \\ \hline
No Cloud & 3.22 $\pm$ 1.52 & 2.44 $\pm$ 1.57 & \textbf{2.35 $\pm$ 1.44} & 2.37 $\pm$ 1.43 \\
\hline
Cirrus & 6.32 $\pm$ 2.91 & 5.65 $\pm$ 2.58 & 5.50 $\pm$ 2.47 & \textbf{5.25 $\pm$ 2.34} \\
\hline
Altostratus & 9.69 $\pm$ 4.56 & 8.66 $\pm$ 4.23 & 8.40 $\pm$ 4.04 & \textbf{8.23 $\pm$ 3.90} \\
\hline
Altocumulus & 9.51 $\pm$ 4.71 & 9.04 $\pm$ 4.74 & \textbf{8.82 $\pm$ 4.73} & 8.83 $\pm$ 4.70 \\
\hline
Stratus & 7.91 $\pm$ 5.02 & 7.59 $\pm$ 5.13 & 7.29 $\pm$ 4.97 & \textbf{7.22 $\pm$ 4.84} \\
\hline
Stratocumulus & 10.88 $\pm$ 4.95 & 10.08 $\pm$ 4.92 & 10.03 $\pm$ 4.86 & \textbf{9.88 $\pm$ 4.87} \\
\hline
Cumulus & 9.94 $\pm$ 6.47 & 9.44 $\pm$ 6.29 & \textbf{9.21 $\pm$ 6.21} & 9.23 $\pm$ 6.27 \\
\hline
Nimbostratus & 14.12 $\pm$ 5.22 & 12.37 $\pm$ 4.93 & \textbf{11.93 $\pm$ 4.78} & 12.07 $\pm$ 4.85 \\
\hline
Deep Convection & 13.03 $\pm$ 4.82 & 12.11 $\pm$ 4.62 & 11.71 $\pm$ 4.51 & \textbf{11.64 $\pm$ 4.37} \\
\hline
\end{tabular}
\end{flushleft}

\vspace{0.5em}
\begin{flushleft}
\begin{tabular}{|L{2.4cm}|C{2.3cm}|C{2.3cm}|C{2.3cm}|C{2.3cm}|}
\hline
 & \multicolumn{4}{c|}{\textbf{Ice Water Content RMSE (g/m\textsuperscript{3})}} \\ \hline
\textbf{Cloud Type} & \textbf{U-Net Baseline} & \textbf{SWinMAE} & \textbf{\begin{tabular}[c]{@{}c@{}}SWinSatMAE\\(no pre-training)\end{tabular}} & \textbf{\begin{tabular}[c]{@{}c@{}}SWinSatMAE\\(our model)\end{tabular}} \\ \hline
No Cloud & 0.005 $\pm$ 0.016 & \textbf{0.004 $\pm$ 0.016} & \textbf{0.004 $\pm$ 0.014} & \textbf{0.004 $\pm$ 0.018} \\
\hline
Cirrus & 0.028 $\pm$ 0.042 & 0.023 $\pm$ 0.030 & 0.023 $\pm$ 0.029 & \textbf{0.021 $\pm$ 0.026} \\
\hline
Altostratus & 0.095 $\pm$ 0.114 & 0.084 $\pm$ 0.097 & 0.082 $\pm$ 0.095 & \textbf{0.078 $\pm$ 0.092} \\
\hline
Altocumulus & 0.026 $\pm$ 0.043 & 0.025 $\pm$ 0.039 & \textbf{0.024 $\pm$ 0.038} & 0.025 $\pm$ 0.039 \\
\hline
Stratus & 0.030 $\pm$ 0.078 & 0.029 $\pm$ 0.072 & 0.027 $\pm$ 0.069 & \textbf{0.026 $\pm$ 0.066} \\
\hline
Stratocumulus & 0.040 $\pm$ 0.115 & 0.037 $\pm$ 0.104 & 0.038 $\pm$ 0.106 & \textbf{0.034 $\pm$ 0.097} \\
\hline
Cumulus & 0.054 $\pm$ 0.107 & \textbf{0.052 $\pm$ 0.102} & 0.053 $\pm$ 0.103 & 0.055 $\pm$ 0.106 \\
\hline
Nimbostratus & 0.258 $\pm$ 0.188 & 0.232 $\pm$ 0.171 & 0.229 $\pm$ 0.172 & \textbf{0.224 $\pm$ 0.159} \\
\hline
Deep Convection & 0.593 $\pm$ 0.254 & 0.544 $\pm$ 0.242 & 0.539 $\pm$ 0.244 & \textbf{0.533 $\pm$ 0.235} \\
\hline
\end{tabular}
\end{flushleft}

\vspace{0.5em}
\begin{flushleft}
\begin{tabular}{|L{2.4cm}|C{2.3cm}|C{2.3cm}|C{2.3cm}|C{2.3cm}|}
\hline
& \multicolumn{4}{c|}{\textbf{Effective Radius RMSE (\textmu m)}} \\ \hline
\textbf{Cloud Type} & \textbf{U-Net Baseline} & \textbf{SWinMAE} & \textbf{\begin{tabular}[c]{@{}c@{}}SWinSatMAE\\(no pre-training)\end{tabular}} & \textbf{\begin{tabular}[c]{@{}c@{}}SWinSatMAE\\(our model)\end{tabular}} \\ \hline
No Cloud & 6.69 $\pm$ 3.04 & 4.70 $\pm$ 3.17 & \textbf{4.52 $\pm$ 2.88} & 5.03 $\pm$ 3.00 \\
\hline
Cirrus & 17.02 $\pm$ 6.35 & 15.85 $\pm$ 5.83 & 15.47 $\pm$ 5.82 & \textbf{14.34 $\pm$ 5.64} \\
\hline
Altostratus & 19.71 $\pm$ 12.36 & 17.48 $\pm$ 11.00 & 17.05 $\pm$ 10.60 & \textbf{16.19 $\pm$ 10.12} \\
\hline
Altocumulus & 17.42 $\pm$ 13.19 & 15.80 $\pm$ 11.63 & 15.71 $\pm$ 11.50 & \textbf{15.68 $\pm$ 11.13} \\
\hline
Stratus & 9.45 $\pm$ 11.13 & 8.82 $\pm$ 10.75 & 8.86 $\pm$ 10.55 & \textbf{8.26 $\pm$ 10.12} \\
\hline
Stratocumulus & 8.75 $\pm$ 13.64 & 7.62 $\pm$ 12.32 & 7.55 $\pm$ 11.89 & \textbf{7.40 $\pm$ 11.68} \\
\hline
Cumulus & 9.38 $\pm$ 13.71 & 7.87 $\pm$ 11.48 & 7.88 $\pm$ 11.18 & \textbf{7.68 $\pm$ 10.94} \\
\hline
Nimbostratus & 23.90 $\pm$ 13.90 & 18.31 $\pm$ 10.10 & 17.77 $\pm$ 9.83 & \textbf{17.26 $\pm$ 9.33} \\
\hline
Deep Convection & 22.82 $\pm$ 9.33 & 19.88 $\pm$ 8.76 & 19.74 $\pm$ 8.74 & \textbf{18.81 $\pm$ 8.00} \\
\hline
\end{tabular}
\end{flushleft}
\end{table}

\clearpage

\begin{figure}[h]
    \centering
    \includegraphics[width=\linewidth]{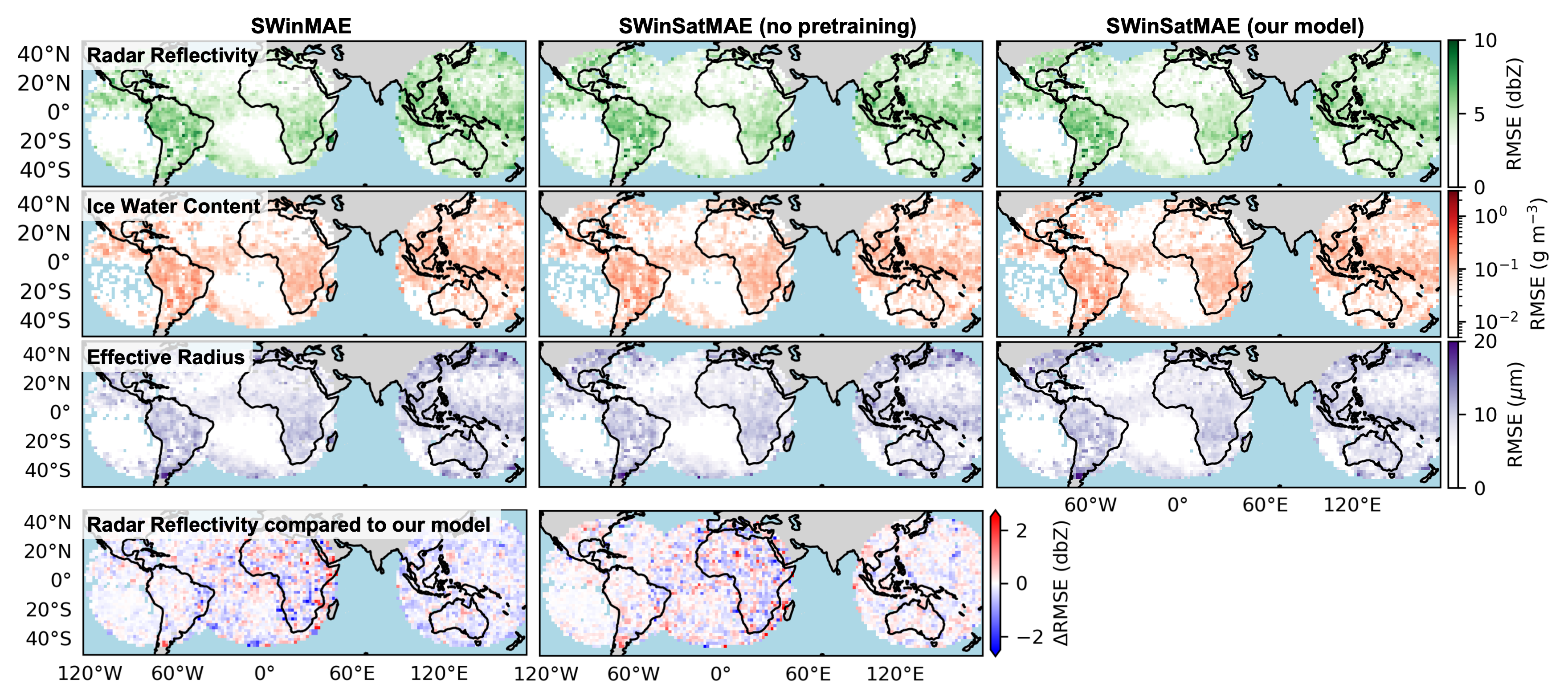}
    \caption{Comparison of three of our models: SWinMAE (no geospatial encodings, left), SWinSatMAE (no pre-training, middle), and our chosen SWinSatMAE architecture that was pretrained on geostationary imagery before fine-tuning to predict CloudSat variables. }
    \label{fig:swinmae_vs_swinsatmae}
\end{figure}

\begin{figure}[h]
    \centering    
    \includegraphics[width=\linewidth]{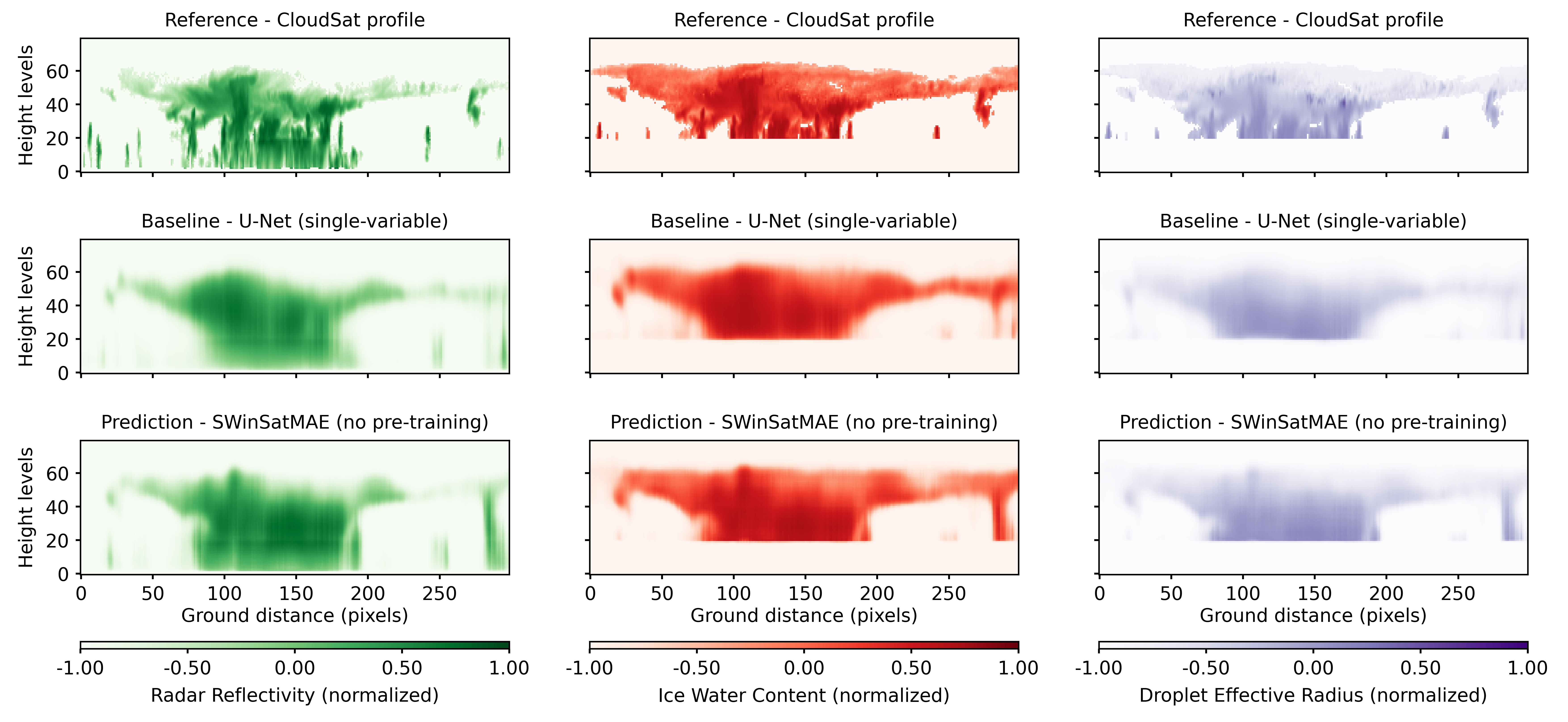}
    \caption{Radar reflectivity (first column), ice water content (second column), and droplet effective radius (third column) as retrieved by CloudSat (first row) along its swath through hurricane Dorian and as reconstructed by the single-variable U-Net baseline (second row) and the SWinSatMAE model without pre-training (third row). We refer the reader to fig. \ref{fig:cross-sections} for the context image of hurricane Dorian as well as the reconstructions by the multi-variable U-Net Baseline and the SWinSatMAE model with pre-training.}
    \label{appendix-figure-models-dorian-slices}
\end{figure}

\begin{figure}[h]
    \centering    \includegraphics[width=0.75\linewidth]{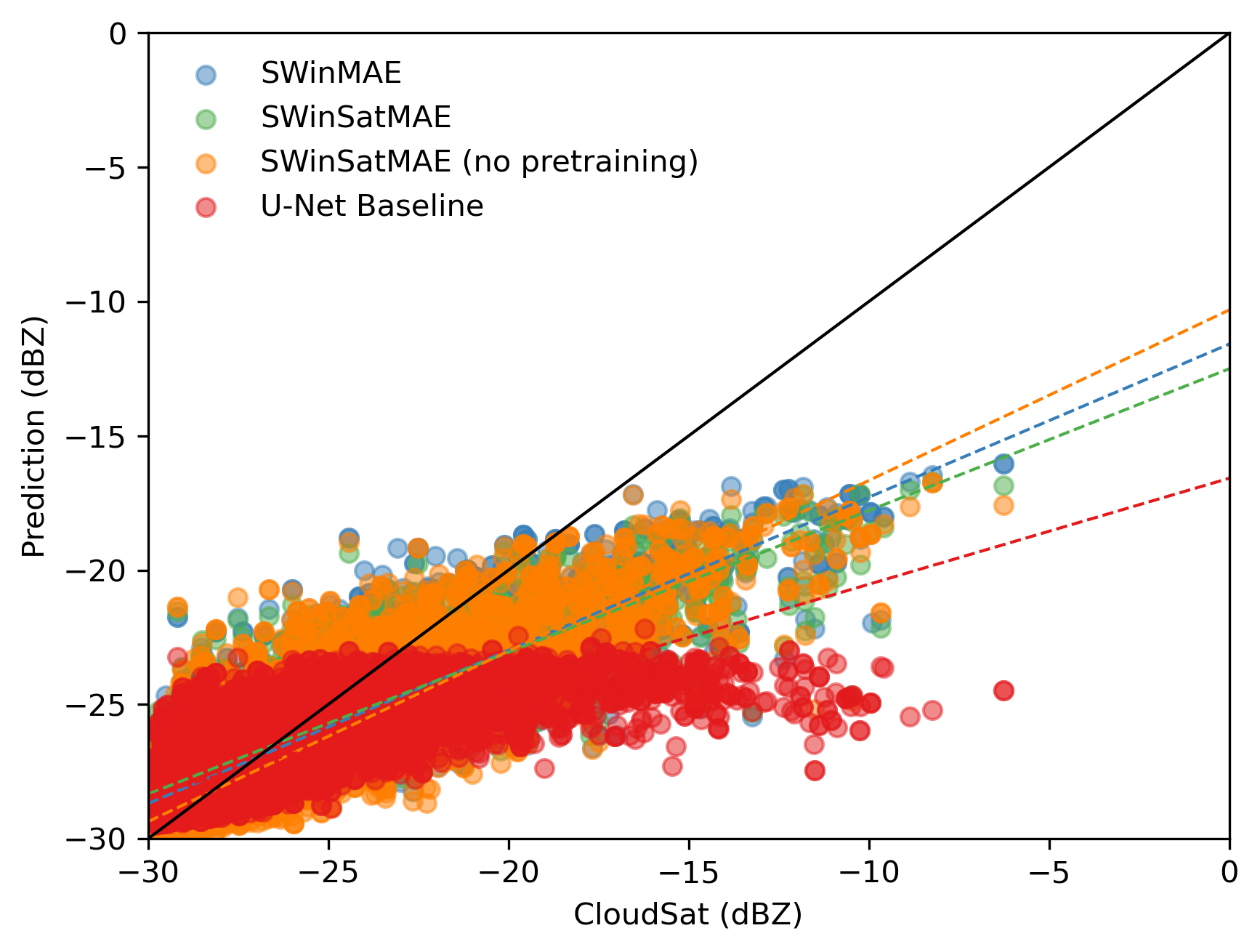}
    \caption{\textbf{Clear \& Cloudy}: Comparison between the CloudSat radar reflectivity (mean per profile) and our model predictions. We consider all pixels, i.e. clear and cloudy for this plot.}
    \label{appendix-figure-target-predictions}
\end{figure}

\begin{figure}[h]
    \centering    \includegraphics[width=\linewidth]{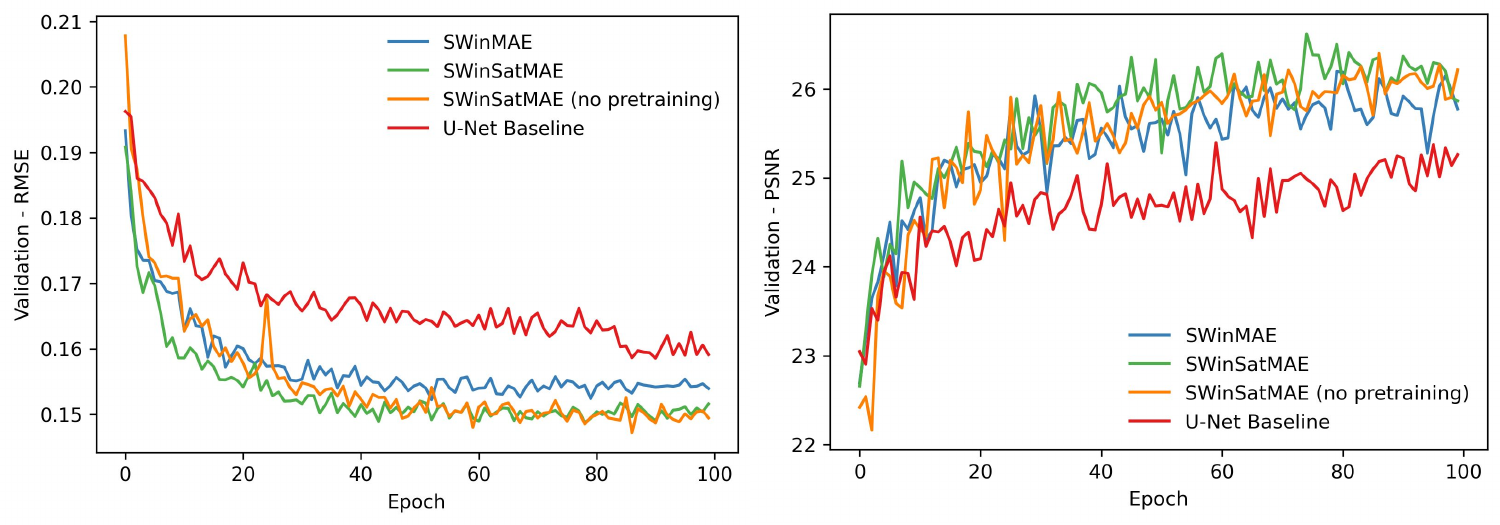}
    \caption{Root-mean-square error (RMSE) and peak-signal-to-noise ratio (PSNR) during training.}
    \label{appendix-figure-training-metrics}
\end{figure}

\clearpage

\subsection*{Cloud and Tropical Cyclone Reconstructions}

\end{document}